\documentclass{article}
\usepackage{array}

\usepackage{graphicx}
\usepackage{amsmath}
\usepackage{amssymb}
\usepackage{booktabs}
\usepackage{mathtools}   % loads »amsmath«
\usepackage{multirow}
\usepackage[table]{xcolor}
\usepackage{graphicx}
\usepackage{subcaption}
\usepackage{booktabs}  

\usepackage{url}
\usepackage{makecell}

\usepackage{times}
\usepackage{epsfig}
\usepackage{graphicx}
\usepackage{amsmath}
\usepackage{amssymb}
\usepackage{fontawesome5}
\usepackage{graphicx}
\usepackage{amsmath}
\usepackage{amssymb}
\usepackage{booktabs}
\usepackage[ruled,vlined]{algorithm2e}
\usepackage{algorithmic}
\usepackage{colortbl} % for coloring cells
\usepackage{booktabs} % for horizontal rules
\usepackage{multirow, booktabs, graphicx, xcolor, colortbl, arydshln, tikz}
\usetikzlibrary{tikzmark}
\usepackage{makecell}
\usepackage{siunitx} % for 'S' column type
\usepackage{caption}
\usepackage{stfloats}

\PassOptionsToPackage{numbers, compress}{natbib}
\usepackage[preprint]{neurips_2025}

\usepackage[utf8]{inputenc} % allow utf-8 input
\usepackage[T1]{fontenc}    % use 8-bit T1 fonts
\usepackage{hyperref}       % hyperlinks
\usepackage{url}            % simple URL typesetting
\usepackage{booktabs}       % professional-quality tables
\usepackage{amsfonts}       % blackboard math symbols
\usepackage{nicefrac}       % compact symbols for 1/2, etc.
\usepackage{microtype}      % microtypography
\usepackage{xcolor}         % colors
\usepackage{graphicx}

\title{Towards Multimodal Understanding via\\Stable Diffusion as a Task-Aware Feature Extractor}

\author{%
  Vatsal Agarwal\thanks{Work done during an internship at Apple} \\
  University of Maryland\\
  \And 
Matthew Gwilliam \\
University of Maryland 
  \And 
  Gefen Kohavi \\
  Apple \\ 
  \And
Eshan Verma \\
  Apple \\ 
  \And 
  Daniel Ulbricht \\ 
  Apple 
  \And 
  Abhinav Shrivastava \\
  University of Maryland
}
\begin{document}

\maketitle

\begin{abstract}

Recent advances in multimodal large language models (MLLMs) have enabled image-based question-answering capabilities. However, a key limitation is the use of CLIP as the visual encoder; while it can capture coarse global information, it often can miss fine-grained details that are relevant to the input query. To address these shortcomings, this work studies whether pre-trained text-to-image diffusion models can serve as instruction-aware visual encoders. Through an analysis of their internal representations, we find diffusion features are both rich in semantics and can encode strong image-text alignment. Moreover, we find that we can leverage text conditioning to focus the model on regions relevant to the input question. We then investigate how to align these features with large language models and uncover a leakage phenomenon, where the LLM can inadvertently recover information from the original diffusion prompt. We analyze the causes of this leakage and propose a mitigation strategy. Based on these insights, we explore a simple fusion strategy that utilizes both CLIP and conditional diffusion features. We evaluate our approach on both general VQA and specialized MLLM benchmarks, demonstrating the promise of diffusion models for visual understanding, particularly in vision-centric tasks that require spatial and compositional reasoning. Our project page can be found \href{https://vatsalag99.github.io/mustafar/}{here}. 

% Our key insight lies in the fact that these models possess rich intermediate representations that are vision-centric and can be guided through text instructions, enabling them to facilitate multimodal understanding. 

% We demonstrate this robust understanding of vision language through the image-text matching task. Subsequently, we elaborate on the key design decisions for repurposing text-to-image generative models as task-aware feature extractors. We highlight instances where these models demonstrate exceptional performance, particularly in spatial and compositional understanding. Finally, we evaluate our approach across a diverse range of both general VQA and more specialized MLLM benchmarks to assess the strengths and limitations of text-to-image models in visual understanding tasks.  

\end{abstract}

\section{Introduction}

\begin{figure*}
    \centering
    \includegraphics[width=\linewidth]{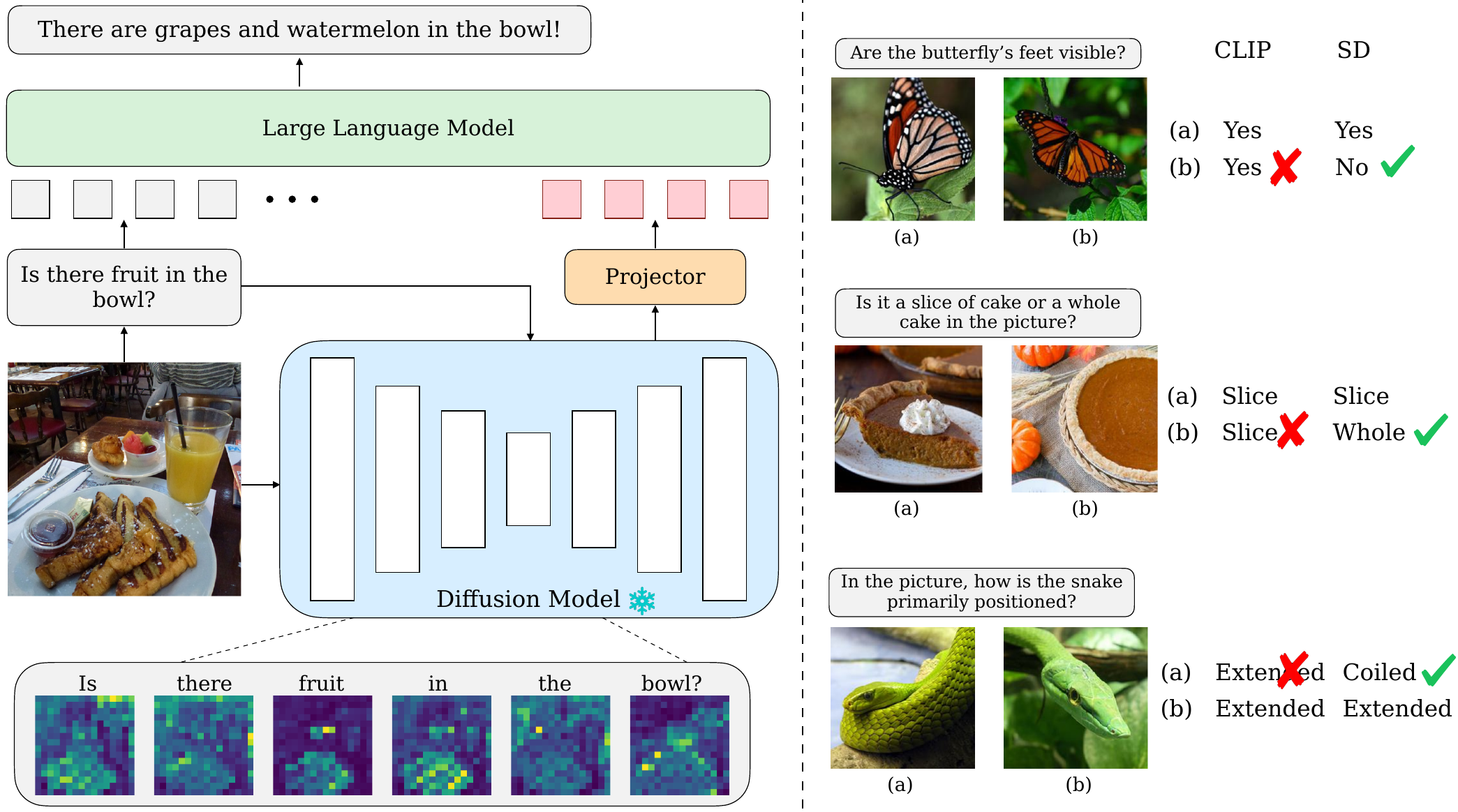}
    \caption{\textbf{Overview.} \textbf{(Left)} We present our full multimodal pipeline. Following LLaVA, we first extract visual features from the frozen diffusion model and pass the question as text-prompt. The LLM then uses these features to generate its answer. Cross-attention maps show that the model can use the question to focus on relevant regions \textbf{(Right)} We show examples on MMVP~\cite{tong2024eyes} where diffusion features outperform CLIP. }
    \label{fig:teaser}
    \vspace{-1em}
\end{figure*}

% \begin{figure}[t]
%   \centering
%   \fbox{\rule{0pt}{2in} \rule{3in}{0pt}} % 2in tall, 3in wide placeholder box
%   \caption{
%     In this work, we investigate how Stable Diffusion captures image-text relationships and whether its internal representations can serve as strong visual encoders for multimodal understanding. We analyze cross-attention maps and find that Stable Diffusion dynamically localizes concepts based on text input, with attention responding to object presence or absence. On the MMVP-VLM benchmark, SD outperforms CLIP in categories like viewpoint, compositionality, and structure, showing stronger performance on fine-grained visual factors. These insights suggest diffusion features offer complementary, task-aware representations and motivate their use for vision-language modeling in downstream tasks like VQA.
%   }
%   \label{fig:teaser}
% \end{figure}

% \begin{figure}[h!]
%     \centering
%     \includegraphics[width=\linewidth]{figures/plot.pdf}
%     \includegraphics[width=\linewidth]{figures/teaser_attn.pdf}
%     \caption{Text-to-image diffusion models such as Stable-Diffusion have strong capabilities for vision-language understanding. Stable-Diffusion outperforms CLIP on the challenging  MMVP-VLM image-text matching benchmark~\cite{tong2024eyes} (shown on top), sorted by the in performance across various visual factors. Stable-Diffusion leverages its internal text-conditioned cross-attention maps (shown at the bottom).}
%     \label{fig:teaser}
%     \vspace{-0.2in}
% \end{figure}

Effective visual feature extraction remains a key challenge in designing robust multimodal large language models (MLLMs). 
Due to its vision-language pre-training, CLIP~\cite{radford2021learning} is the de facto visual encoder for MLLM architectures, despite its inability to encode fine-grained visual details~\cite{tong2024eyes, monsefi2024detailclip, bianchi2024clipmainroadblockfinegrained} and capture compositional information~\cite{he2023discffusion, bianchi2024clipmainroadblockfinegrained}. 
%However, its visual representations often fail to encode fine-grained visual details such as orientation and structure~\cite{tong2024eyes, monsefi2024detailclip, bianchi2024clipmainroadblockfinegrained}. 
%Furthermore, they are limited in how well they capture compositional information, such as the binding of objects and actions or objects and attributes~\cite{he2023discffusion, bianchi2024clipmainroadblockfinegrained}.
%To address these shortcomings, recent works ensemble multiple visual encoders~\cite{kar2024brave, tong2024cambrian, tong2024eyes, jiang2023from, li2024minigeminiminingpotentialmultimodality} or integrate auxiliary modalities such as segmentation or depth~\cite{liu2023prismer, jain2024vcoder}. 
%However, these approaches introduce additional computational overhead and require extensive tuning~\cite{yang2024law}.
We contend that recent attempts to address these shortcomings~\cite{kar2024brave, tong2024cambrian, tong2024eyes, jiang2023from, li2024minigeminiminingpotentialmultimodality,liu2023prismer, jain2024vcoder} are flawed in their reliance on static visual features, which may not capture task-relevant visual details.
For instance, answering a question about the visibility of a butterfly's feet, as in Fig.~\ref{fig:teaser} (right), requires precise localization and fine-grained semantic understanding.
%, a process that mirrors how humans dynamically attend to visual input~\cite{treisman1980feature, wolfe2004attributes}.
Although recent methods explore question-aware visual representations~\cite{ganz2024questionawarevisiontransformer, yu2024api}, they often require substantial architectural changes to CLIP and employ fusion at a single stage thereby limiting image-query interactions~\cite{ganz2024questionawarevisiontransformer, instructblip, li2023blip}.

In this work, we leverage text-to-image diffusion models as the visual encoder (pipeline shown in Fig.~\ref{fig:teaser} (left)). 
These generative networks can create high-quality images that align well with the semantics and compositions described by a given text prompt~\cite{rombach2022high,NEURIPS2022_ec795aea,ramesh2022hierarchicaltextconditionalimagegeneration,podell2023sdxlimprovinglatentdiffusion}.
Recent studies find that this is a result of their strong internal representations as well as the cross-attention mechanism, which is embedded throughout their architecture and modulates pixel features with the input text~\cite{hertz2022prompt, tang2022daaminterpretingstablediffusion, liu2024understandingcrossselfattentionstable, tumanyan2023plug}. 

We first analyze unconditional features extracted from diffusion models across multiple blocks and timesteps. 
Although prior work has shown that such features can be repurposed for tasks like classification, segmentation, and depth estimation~\cite{wang2023diffusion, ke2023repurposing, mukhopadhyay2023text, chen2024deconstructingdenoisingdiffusionmodels}, their utility for multimodal reasoning remains underexplored. 
%Thus, we perform a detailed investigation of diffusion features to identify their intrinsic strengths for fine-grained visual understanding. 
%Specifically, we extract off-the-shelf unconditional diffusion features across multiple blocks and timesteps. 
We use the LLaVA framework for our investigation and inspect model performance using features across various blocks and timesteps. 
Our results on general-purpose and vision-centric benchmarks show that these features encode complementary information and, in many cases, match or outperform CLIP-based LLaVA models. 
We show examples in Fig~\ref{fig:teaser} (right). 

Building on this, we inspect the text-conditional cross-attention maps.
%that are embedded throughout the diffusion architecture and modulate pixel-level features using the input text~\cite{hertz2022prompt, tang2022daaminterpretingstablediffusion, liu2024understandingcrossselfattentionstable}. 
Fig.~\ref{fig:teaser} (left) shows that these attention maps can focus on semantically relevant regions. 
%To quantify this alignment, we evaluate image-text matching performance on Winoground~\cite{thrush_and_ross2022winoground} and MMVP-VLM~\cite{tong2024eyes} and observe that diffusion cross-attention significantly outperforms CLIP and its variants in capturing spatial and compositional relationships (see Table~\ref{tab:mmvp_winoground_vlm}). 
We seek to quantify this alignment via image-text matching performance and find that diffusion cross-attention significantly outperforms CLIP in capturing spatial and compositional relationships (see Table~\ref{tab:mmvp_winoground_vlm}).
We also investigate how text conditioning modulates intermediate spatial features at different blocks and layers. 
%Our findings demonstrate that providing text leads to small modulations with minimal structural changes. 
%However, the strength of impact is different for varying blocks and layers. 
%Additionally, we examine how text guidance impacts the alignment between diffusion features and the language space. 
%Here, we use image captioning as a proxy task and follow the LLaVA pre-training protocol, where the vision encoder and LLM are frozen and only a projection layer is learned. 
%To measure alignment, we evaluate the model using the COCO-captions benchmark~\cite{DBLP:journals/corr/ChenFLVGDZ15, karpathy2015deep}. 
%We find that while SD2.1 is unable to align to the language-space as well as CLIP, providing captions as text input can greatly reduce this gap. 
%We assess how amplifying the effect of text impacts alignment and provide notable findings in this setting. 
While the impact is small for lower guidance values, amplifying this guidance can visibly change the structure of the features, with greater focus placed on image regions related to the text prompt. 
Furthermore, we discover a leakage effect where the language model can learn to extract the input caption fed to the diffusion model. 
We show how  to quantify the effect of leakage and provide a mitigation strategy during training. 

We then turn our focus towards using questions as the text condition for the diffusion model. 
First, we demonstrate that using questions as text-input guides the intermediate features to highlight relevant regions. 
Based on our findings, we propose leveraging the complementary information in CLIP and diffusion features to build an improved multimodal pipeline and show encouraging performance.

In summary, our contributions are as follows:

\begin{itemize}

%\item{We probe the structure and semantics in unconditional diffusion features and show that they contain fine-grained knowledge leading to improved performance on vision-centric tasks with +1.23\% on BLINK-val.}
\item{We show that unconditional diffusion features contain fine-grained structure and semantics, leading to improved performance on vision-centric tasks with +1.23\% on BLINK-val.}

\item{We investigate what information is stored in text-conditioned cross-attention maps and find that they encode robust vision-language correspondence. We show that when used for image-text matching, these maps can match or exceed CLIP with a +5\% gain on MMVP-VLM. }

\item{We provide comprehensive analysis to show how text conditioning modulates spatial features, enabling greater focus on query-specific regions.}

\item We explore harnessing the complementary information in CLIP and conditional diffusion features and achieve promising results with a +6\% improvement on MMVP over the LLaVA-v1.5-7B baseline.

\end{itemize}

\begin{figure}[tbp]
\centering
\includegraphics[width=\linewidth]{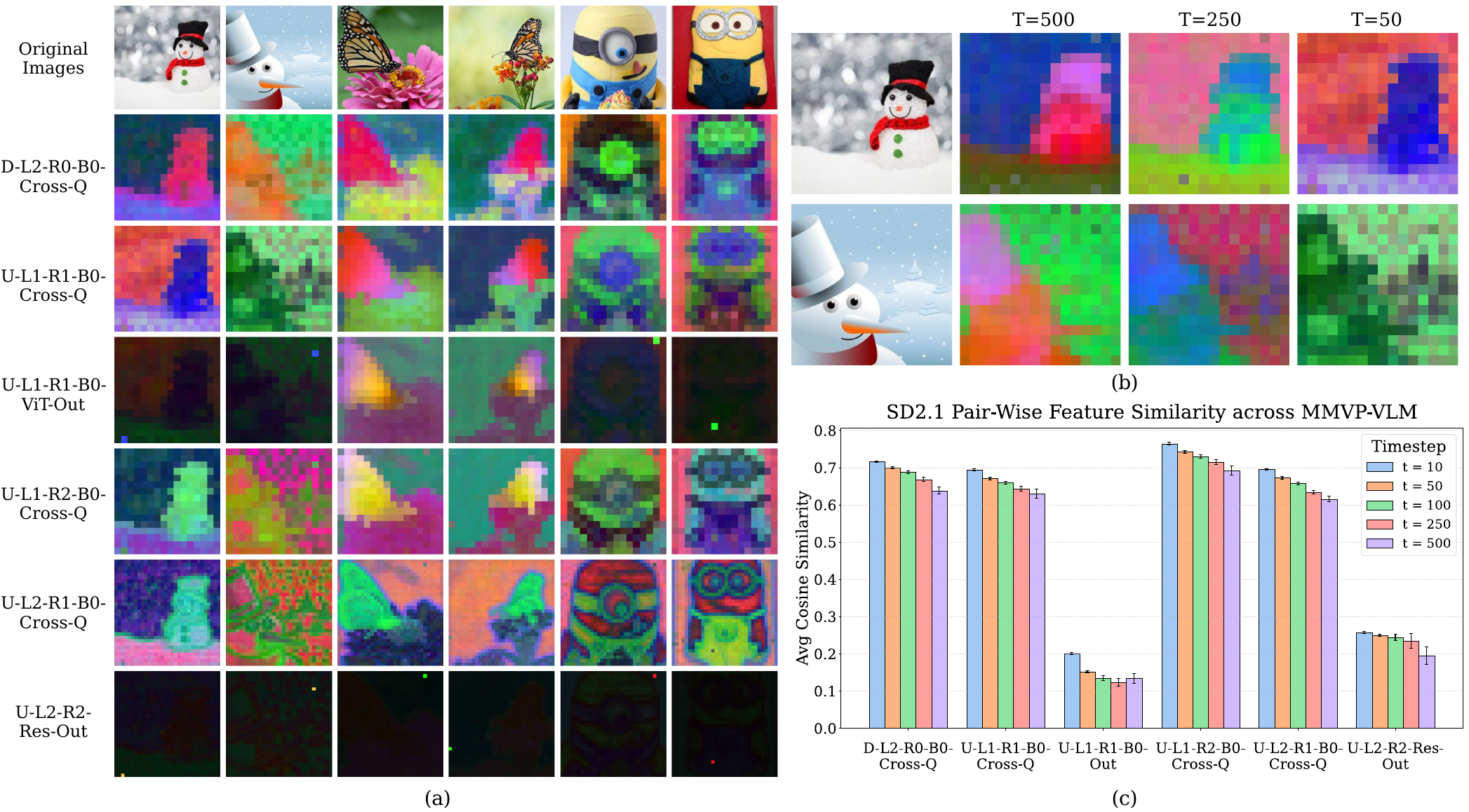}
\caption{\textbf{Inspecting Diffusion Features}. (a, b) We visualize spatial features for three image pairs from MMVP-VLM using PCA across blocks and timesteps. For (a), we use T=50; for (b), we fix \texttt{U-L1-R1-B0-Cross-Q}. We observe: (1) different blocks capture either shared semantics or image-specific details; (2) higher timesteps encode coarse layout, while lower timesteps emphasize fine-grained structure; and (3) features like \texttt{out} and \texttt{res-out} have tokens which can act as "registers" that act as shared global descriptors across similar images. (c) We plot average cosine similarity, finding that \texttt{cross-q} representations capture greater similarity compared to \texttt{out} features.}
\label{fig:pca_analysis_diff_features_uncond}
\end{figure} 

\section{Related Work}
\label{sec:rel_work}

\textbf{Vision Language Models} Vision language modeling has been a popular topic with foundational image-text alignment papers such as~\cite{Yu2022CoCaCC, radford2021learning}. Multimodal LLMs go one step further and have taken the success of large-scale pretrained LLMs and applied them to vision tasks~\cite{neurips_frozen,neurips_flamingo, li2023blip, instructblip}. However, they typically require a large amount of pre- or post-training to align the vision and language features. More recent methods like~\cite{liu2023llava, zhu2023minigpt, chen2023minigptv2} show how visual instruction can be done quickly with low data while being competitive with strong baselines~\cite{awadalla2023openflamingo, li2023blip} across a wide variety of tasks.

Numerous works and benchmarks have demonstrated the shortcomings of these methods, including their propensity for hallucination~\cite{li2023evaluating, favero2024multi, gunjal2024detecting, huang2023survey, zhai2023halle, sun2023aligning}. Furthermore, these methods exhibit a general inability to perform spatial reasoning tasks~\cite{fu2024blink, tong2024cambrian, tong2024eyes}. Several improvements have been proposed, such as increasing the resolution~\cite{liu2024llavanext, mckinzie2024mm1}, enhancing data mixtures~\cite{liu2023improvedllava, tong2024cambrian}, and combining or swapping with other encoders~\cite{li2024minigeminiminingpotentialmultimodality, tong2024eyes, tong2024cambrian}. We demonstrate that further research is necessary to enhance vision encoders and their ability to be prompt-aware.

\noindent\textbf{Combining Visual Features with Other Modalities}
Additional modalities have been proven useful for language tasks.~\cite{mizrahi20244m, bachmann20244m} show how a masking objective can connect more modes than RGB to language.~\cite{liu2023llava, liu2023llavaplus, gupta2023visual} show how tool use alongside extra modalities can significantly expand use cases. Papers such as~\cite{jain2024vcoder, liu2023prismer, cai2024spatialbotprecisespatialunderstanding} show how integrating extra modalities such as depth and semantic segmentation helps improve results on topics such as counting and spatial reasoning. Despite these additions, none of these models are aware of visual instruction input and therefore cannot focus on features that maximize performance for a single prompt.

\noindent\textbf{Diffusion Models for Discriminative Tasks} There have been multiple works that look at porting diffusion models from generative tasks to discriminative tasks. The Diffusion Classifier~\cite{li2023diffusion}  shows how to rework a standard class-conditional diffusion model into a discriminative classifier.~\cite{mukhopadhyay2023text} identifies where and when in a diffusion U-Net provides the strongest discriminative features.

\begin{figure}[h]
\centering
\begin{subfigure}[b]{0.42\textwidth}
    \centering
    \tiny % Using smaller font size (scriptsize instead of footnotesize)
    \begin{tabular}{@{}l@{\hspace{1mm}}c@{\hspace{1mm}}c@{\hspace{1mm}}c@{\hspace{1mm}}c@{\hspace{1mm}}c@{\hspace{1mm}}c@{}}
    \toprule
    \multirow{2}{*}{\textbf{Config}} & \textbf{LLaVA-B} & \textbf{MMVP} & \textbf{BLINK-val} & \multicolumn{3}{c}{\textbf{Natural-Bench}} \\
    \cmidrule(r){2-2} \cmidrule(r){3-3} \cmidrule(r){4-4} \cmidrule(){5-7}
    & All & Acc & Acc & Q-Acc & I-Acc & G-Acc \\
    \midrule
    LLaVA-v1.5-7B & 66.5 & 24.7 & 36.60 & 37.70 & 43.80 & 14.32 \\
    \midrule
    \multicolumn{6}{@{}l}{\textit{Layer Configurations at T = 50}} \\
    \midrule
    \texttt{D-L2-R0-B0-Cross-Q} & 33.4 & 17.3 & 35.81 & 26.71 & 34.87 & 7.11 \\
    \texttt{U-L1-R1-B0-Cross-Q} &\textbf{ 45.3} & \textbf{22.7} & 37.28 &  30.23 & 37.05 & \textbf{9.11} \\
    \texttt{U-L1-R1-B0-Out} & 41.1 & 22.0 & \textbf{37.83} & 28.82 & 37.05 & 8.68 \\
    \texttt{U-L1-R2-B0-Cross-Q} & 42.2 & \textbf{22.7} & 34.87 & \textbf{31.13} & \textbf{37.87} & 8.42 \\
    \texttt{U-L2-R1-B0-Cross-Q} & 39.0 & 22.0 & 37.23 & 28.82 & 35.71 & 8.95 \\
    \texttt{U-L2-R2-Res-Out} & 34.2 & 19.3 & 35.65 & 27.68 & 35.68 & 6.74 \\
    \midrule
    \multicolumn{6}{@{}l}{\textit{Time-Steps for \texttt{U-L1-R1-B0-Cross-Q}}} \\
    \midrule
    T = 10 & 40.0 & 22.0 & 36.82 & 30.29 & 36.61 & 8.89 \\
    T = 50 & \textbf{45.3} & \textbf{22.7} & 37.28 & 30.23 & 37.05 & 9.11 \\
    T = 100 & 44.7 & 22.0 & 37.37 & \textbf{30.87} & \textbf{38.45} & \textbf{9.42} \\
    T = 250 & 41.3 & 18.0 & \textbf{37.39} & 26.82 & 34.50 & 7.11 \\
    T = 500 & 29.8 & 12.7 & 36.54 & 16.87 & 27.61 & 3.11 \\
    \bottomrule
    \end{tabular}
\end{subfigure}
\hfill
\begin{subfigure}[b]{0.48\textwidth}
    \centering
    \includegraphics[width=\textwidth]{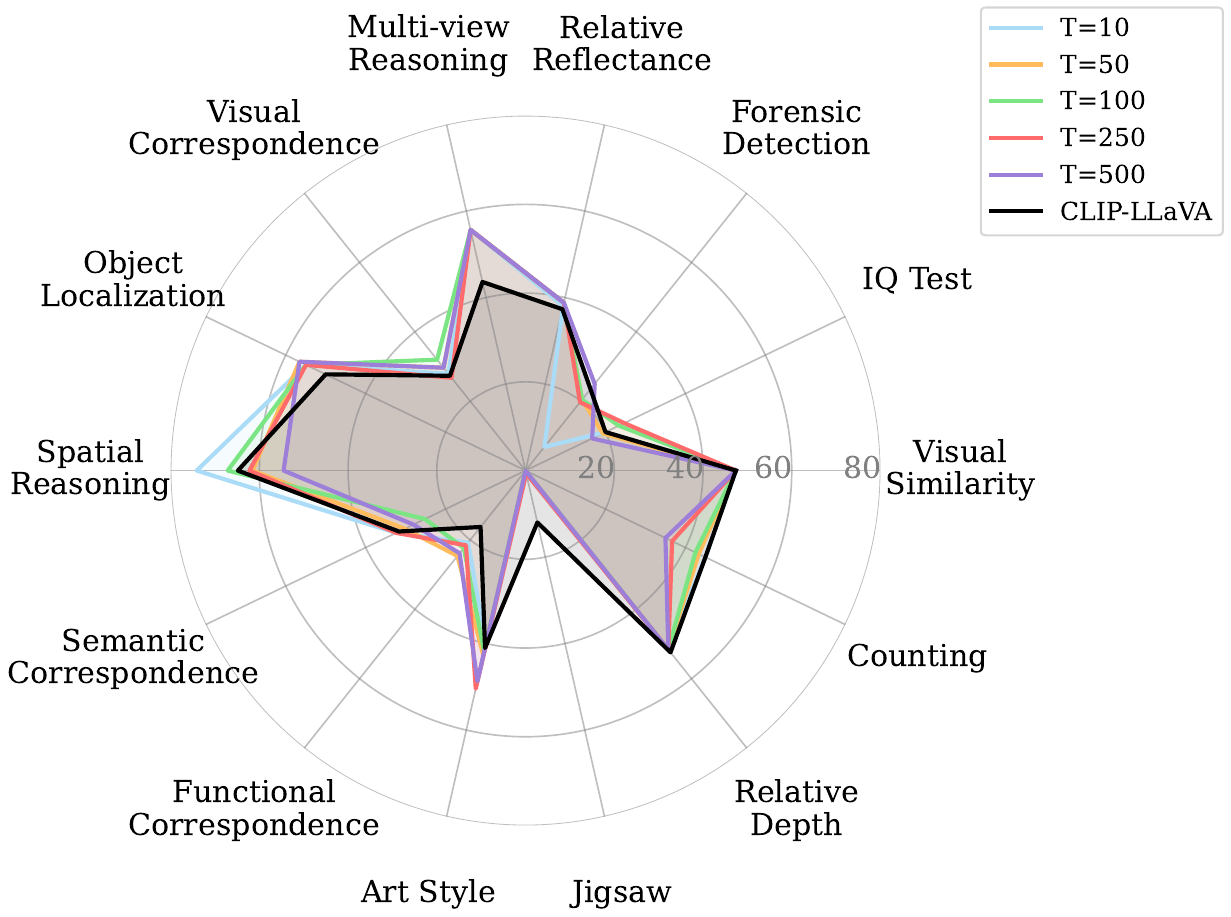}
    \vspace{-80pt}
\end{subfigure}
\vspace{1em}
\caption{\textbf{General Model Performance (Left):} We evaluate multimodal reasoning using the LLaVA framework with diffusion features at different layers and timesteps. The table reports accuracy on LLaVA-Bench, MMVP, and NaturalBench under varying feature extraction points. \textbf{BLINK-val Performance (Right):} The plot shows BLINK-val benchmark performance across different timesteps. SD-based models consistently outperform CLIP (in black) across timesteps. We point out notable improvements: +10\% on Spatial Reasoning, Multi-view Reasoning, and Art Style.}
\label{fig:model_performance_diff_features_uncond}
\end{figure}

These discriminative features are useful for multiple tasks. For classification,~\cite{mukhopadhyay2023diffusion} explores feature extraction and shows how diffusion models are stronger than other generative models on discriminative tasks. There has also been significant exploration on using diffusion models as encoders for segmentation tasks~\cite{xu2023odise, karazija2023diffusion}. Particularly~\cite{xu2023odise} shows how diffusion models have both strong open vocabulary and region-level understanding by achieving SoTA performance using a frozen diffusion backbone. Finally,~\cite{he2023discffusion} explains that diffusion models can achieve state-of-the-art on few-shot image-text matching. Following a similar strategy to these papers, we show how diffusion models can provide sufficiently strong discriminative features for visual instruction tuning.

\section{What Visual Information Do Diffusion Models Encode?}
\label{sec:uncond}

In this section, we investigate the quality of diffusion representations for visual question answering. We aim to understand how well unconditioned diffusion features align with the language space of an LLM and quantify the performance of features extracted at different blocks and timesteps. In this work, we use Stable Diffusion
v2.1-base~\cite{rombach2022high} as our diffusion model as it has been trained on a large and diverse set of image pairs and has shown excellent generative capabilities. 

\subsection{Diffusion Features Encode Semantic and Structural Information}
\label{subsec:uncond_analysis}

We first analyze intermediate diffusion features via PCA building off the codebase from ~\cite{meng2024diffusionmodelactivationsevaluated}. We aim to understand how features across layers and timesteps encode semantic and structural information. For each block and selected timestep, we extract pixel-wise features and project them onto their top three principal components, allowing us to visualize spatial patterns. These visualizations qualitatively reveal how features represent spatial layout and capture both coarse- and fine-grained structural details. We present the results in Fig.~\ref{fig:pca_analysis_diff_features_uncond}, and we highlight key findings below.

We primarily inspect features used in cross-attention layers, specifically the pixel-wise queries, as they most directly interact with text embeddings and are likely to be more semantically aligned. To complement this, we also examine features extracted after the cross-attention operation (\texttt{b0-out}) and after the full residual attention block (\texttt{res-out}). We use a subset of image pairs from the MMVP-VLM dataset, extracting features at timestep 50 when the noise level is relatively low.

\begin{figure}
    \centering
    \includegraphics[width=1\linewidth]{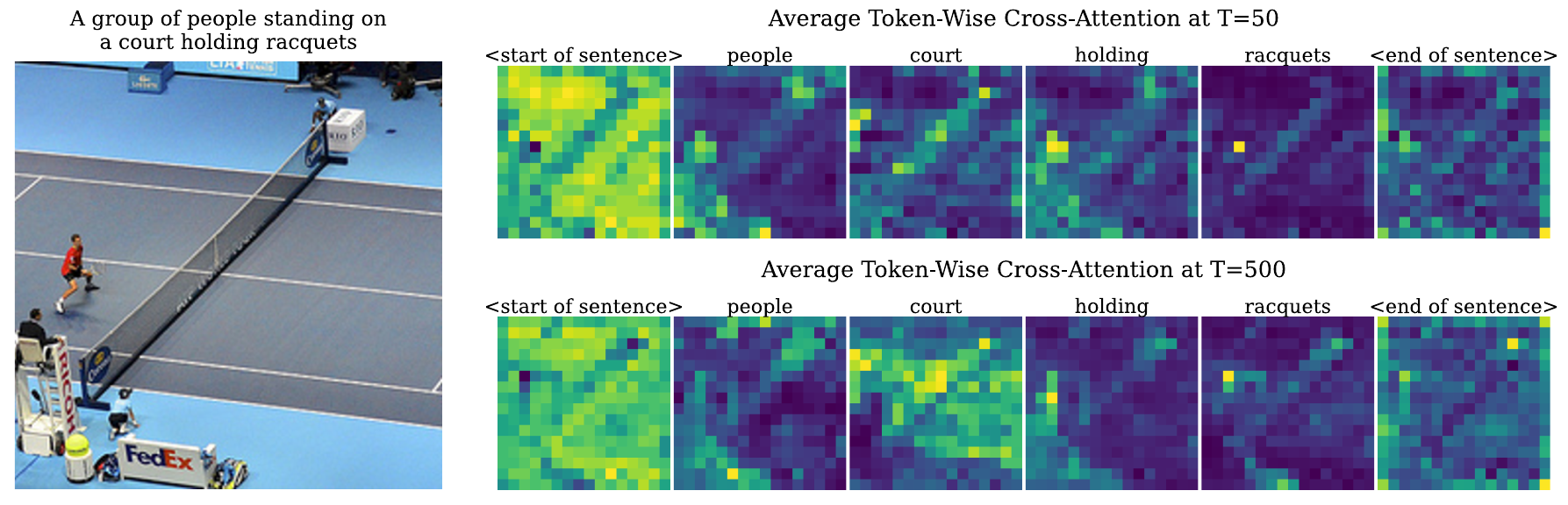}
    \caption{\textbf{Visualizing Cross-Attention Maps.} We show a sample from COCO-captions and averaged cross-attention maps at low and high timesteps for a few key words, representing the attention between pixel features and a specific word. We observe that cross-attention maps at higher timesteps show higher focus on background elements (e.g., ``court''). Attention maps from lower timesteps provide improved localization of both object and action concepts (e.g., ``racquets'' and ``holding'').}
    
    % The \texttt{<start of sentence>} token can act as an attention sink receiving the most attention, while the \texttt{<end of sentence>} token focuses on the overall scene. More examples are provided in the appendix. }
    \label{fig:cross_attn_maps_vis}
\end{figure}

Examining Fig.~\ref{fig:pca_analysis_diff_features_uncond}(a), we observe that these features encode both shared and sample-specific patterns across images. For example, in the pair of minions, PCA maps reveal consistent representations for object components, such as glasses and clothing. These features also capture fine-grained details, as seen in the pair of butterfly images, where the body and wing regions are clearly separated. Interestingly, features extracted after the cross-attention operation (rows 4 and 7) often encode features similar to ``register'' tokens: 1-2 tokens that capture the most variance in the image and are shared between samples. In Fig.~\ref{fig:pca_analysis_diff_features_uncond}(b), we examine how features evolve with timestep. Specifically, we extract features in a single forward pass with noise conditioned on a single timestep. 
% Thus, extracting features at timestep 50 is computationally equivalent to timestep 500. 
We find that higher timesteps capture coarse spatial structure (e.g., the snowman's body and hat), while intermediate timesteps increasingly highlight finer details, such as the scarf (row 1, timestep 250).

Prior work by~\cite{tong2024eyes} identified a key limitation of CLIP: semantically similar yet visually distinct images are often embedded similarly. Motivated by this, we assess whether diffusion features better capture such visual differences. We use the MMVP-VLM benchmark, which was created by including pairs of images that had CLIP similarity over 0.95 and DINO~\cite{oquab2024dinov2learningrobustvisual} similarity below 0.6. We compute cosine similarity between feature maps of similar image pairs across blocks and timesteps as shown in Fig.~\ref{fig:pca_analysis_diff_features_uncond}(c). Several notable trends can be observed: (1) diffusion features capture intra-pair visual differences better compared to CLIP; (2) \texttt{cross-q} features show consistently higher pairwise similarity than \texttt{b0-out} and \texttt{res-out} features, which aligns with our earlier observations that output features encode more image-specific content; and (3) pairwise similarity decreases as the timestep increases. We hypothesize this is a result of the increased noise causing the diffusion features to encode more visually distinct features. 

\subsection{Diffusion Features are Effective for Vision-Centric Tasks}
\label{subsec:diff_uncond}

We investigate the effectiveness of diffusion features on multimodal understanding tasks, adopting the LLaVA architecture~\cite{liu2023llava, liu2023improvedllava} as our training framework. We begin by using off-the-shelf diffusion features and train a two-layer MLP projection head on the LLaVA-558k subset for pretraining~\cite{liu2023improvedllava}. This is followed by fine-tuning both the projection head and the language model (Vicuna-7B~\cite{chiang2023vicuna}) on the LLaVA-Mix-665k SFT dataset~\cite{liu2023improvedllava}. All training is conducted on 4 H100 GPUs, requiring approximately 4 hours for pretraining and 10 hours for fine-tuning. We defer readers to~\cite{liu2023improvedllava} for all training hyperparameters. We inspect performance across various blocks and timesteps. We evaluate using a two sets of benchmarks broadly covering instruction-following (free form generation) and visual perception (multiple choice). LLaVA-Bench-In-the-Wild (LLaVA-B) assesses instruction understanding on 24 out-of-distribution images with 60 questions requiring world knowledge. For core visual reasoning, we use MMVP~\cite{tong2024eyes} and NaturalBench~\cite{naturalbench} which focus on spatial reasoning (e.g.orientation, color, presence of specific features). Additionally we also use BLINK~\cite{fu2024blink} which benchmarks vision-centric reasoning via visual prompting.

% We inspect performance across various blocks and timesteps. We evaluate using a diverse set of benchmarks targeting both instruction-following and visual perception. LLaVA-Bench-In-the-Wild (LLaVA-B) assesses instruction understanding on 24 out-of-distribution images with 60 questions requiring world knowledge. For core visual reasoning, we use MMVP~\cite{tong2024eyes}, NaturalBench~\cite{naturalbench}, and BLINK~\cite{fu2024blink}. MMVP and NaturalBench focus on spatial reasoning (e.g.orientation, color, presence of specific features), while BLINK extends this to multi-image understanding, including semantic alignment and relative depth. Notably, LLaVA-Bench uses free-form generation, whereas the vision benchmarks are multiple choice.

We show our results in Fig~\ref{fig:model_performance_diff_features_uncond}. On the left, we show model performance on all benchmarks across blocks and timesteps, and on the right, we examine more granular task-level performance on the BLINK-val benchmark. First, we observe that on LLaVA-Bench, the baseline CLIP-based LLaVA model performs the best while SD-based models have almost a 20 point degradation. For other benchmarks, we observe less degradation as SD-based models approach baseline performance on MMVP with a 2 point gap. On BLINK-val, SD-based models consistently outperform CLIP models with about a 1.23\% improvement. This illustrates that these features improve on pure vision-centric reasoning. On NaturalBench, we find that CLIP is able to significantly outperform SD-based models potentially due to its better vision-language alignment.

Across blocks, we observe that \texttt{cross-q} features generally perform better than \texttt{out} features such as on LLaVA-Bench and MMVP suggesting an improved alignment to text at these layers. However, we see that the feature extracted from the \texttt{down-stage} performs much worse with a 12 point drop compared to its \texttt{up-stage} equivalent on LLaVA-Bench. This aligns with findings from previous works~\cite{meng2024diffusionmodelactivationsevaluated}, that features at the \texttt{down-stage} contain more diffusion noise. Looking at timesteps, we observe that features extracted at earlier timesteps between timestep 10-100 perform well compared to earlier or later timesteps. We hypothesize this is due to features from later timesteps losing too many fine-grained details. Inspecting BLINK-val performance (right of Fig.~\ref{fig:model_performance_diff_features_uncond}), we observe that features from different timesteps excel at different tasks. Notably, features at the earliest timestep (t=10) improve over CLIP by 10 points on spatial reasoning and match CLIP on counting. Additionally, diffusion features consistently exceed CLIP performance on multi-view reasoning regardless of timestep. Thus, while diffusion features lag behind CLIP on instruction/knowledge-heavy tasks, they offer strong advantages in pure visual reasoning, particularly when extracted from well-aligned blocks and early-to-mid timesteps.

\begin{table*}[t]
    \centering
    \small % Reducing font size to fit table
    \setlength\tabcolsep{5pt} % Default value: 6pt
    \renewcommand{\arraystretch}{1.1}
    \caption{\textbf{Quantifying the Quality of Diffusion Cross-Attention.}
    We evaluate how well diffusion cross-attention maps capture text-image alignment using the MMVP-VLM and Winoground benchmarks. We compare Stable Diffusion with various CLIP-based models across visual patterns such as viewpoint, structure, and object presence. Icons are used to denote pattern categories:
    \textbf{\faCompass}: Orientation and Direction, \textbf{\faSearch}: Specific Features, \textbf{\faSync}: State and Condition, \textbf{\faSortNumericUp}: Quantity and Count, \textbf{\faMapPin}: Spatial Relations, \textbf{\faPalette}: Appearance, \textbf{\faCogs}: Structure, \textbf{\faFont}: Text, \textbf{\faCamera}: Viewpoint.
    Formatting follows~\cite{tong2024eyes}.}
    
    \vspace{0.1em}
        
    \resizebox{1\textwidth}{!}{
        \begin{tabular}{>{\kern-\tabcolsep}l:ccc:cccccccccc:ccc<{\kern-\tabcolsep}}
        \toprule
        \multicolumn{4}{@{}l@{}|}{\textbf{Model Details}} & 
        \multicolumn{10}{c@{}|}{\textbf{MMVP Visual Patterns}} & 
        \multicolumn{3}{c@{}}{\textbf{Winoground Benchmark}} \\
        \cmidrule{1-4} \cmidrule{5-14} \cmidrule{15-17}
        Model & \shortstack{Image\\Size} & \shortstack{Params\\(M)} & \shortstack{IN-1k\\ZeroShot} &
        \faCompass & \faSearch & \faSync & \faSortNumericUp & \faMapPin & \faPalette & \faCogs & \faFont & \faCamera & \shortstack{Avg.} &
        \shortstack{Text} & \shortstack{Image} & \shortstack{Group} \\
        \midrule
        
        \rowcolor{gray!15} OpenAI ViT-L-14~\citep{radford2021learning} & 224$^2$ & 427.6 & 75.5 & 13.3 & 13.3 & 20.0 & 20.0 & 13.3 & 53.3 & 20.0 & 6.7 & 13.3 & 19.3 & 27.75 &	7.75 &	11.75 \\
        OpenAI ViT-L-14~\citep{radford2021learning}& 336$^2$ & 427.9 & 76.6 & 0.0 & 20.0 & 40.0 & 20.0 & 6.7 & 20.0 & 33.3 & 6.7 & 33.3 &  20.0 & 28.50 &	8.25 &	11.25 \\
        
        OpenCLIP ViT-H-14~\citep{ilharco_gabriel_2021_5143773}& 224$^2$ & 986.1 & 78.0 & 20.0 & 13.3 & 60.0 & 33.3 & 13.3 & 53.3 & 40.0 & 6.7 & 26.7 & 29.6 & 30.68 & 11.91 & 8.36 \\
        
        \rowcolor{gray!15} SigLIP ViT-SO-14~\citep{zhai2023sigmoidlosslanguageimage}& 224$^2$ & 877.4 & 82.0 & 26.7 & 20.0 & 53.3 & \textbf{40.0} & 20.0 & \textbf{66.7} & 40.0 & 20.0 & \textbf{53.3} & 37.8 & 11.75 & 1.25 &	6.50 \\
        SigLIP ViT-SO-14~\citep{zhai2023sigmoidlosslanguageimage}& 384$^2$ & 878.0 & 83.1 &20.0 & 26.7 & 60.0 & 33.3 & 13.3 & \textbf{66.7} & 33.3 & 26.7 & \textbf{53.3} & 37.0 & 17.50 &	4.25 &	11.00 \\
        \rowcolor{gray!15} DFN ViT-H-14~\citep{fang2023datafilteringnetworks}& 224$^2$ & 986.1 & 83.4 & 20.0 & 26.7 & \textbf{73.3} & 26.7 & 26.7 & \textbf{66.7} & \textbf{46.7} & 13.3 & \textbf{53.3} & \textbf{39.3} & \textbf{38.50} &	11.50 &	14.25 \\
        DFN ViT-H-14~\citep{fang2023datafilteringnetworks}& 378$^2$ & 986.7 & \textbf{84.4} & 13.3 & 20.0 & 53.3 & 33.3 & 26.7 & \textbf{66.7} & 40.0 & 20.0 & 40.0 & 34.8 &  \textbf{38.50} & 13.25 &	\textbf{15.25} \\ 
        \rowcolor{gray!15} MetaCLIP ViT-L-14~\citep{xu2024demystifyingclipdata}& 224$^2$ & 427.6 & 79.2 & 13.3 & 6.7 & 66.7 & 6.7 & \textbf{33.3} & 46.7 & 20.0 & 6.7 & 13.3 & 23.7 & 32.50 &	10.75 &	\textbf{15.25}  \\
        MetaCLIP ViT-H-14~\citep{xu2024demystifyingclipdata}& 224$^2$ & 986.1 & 80.6 & 6.7 & 13.3 & 60.0 & 13.3 & 6.7 & 53.3 & 26.7 & 13.3 & 33.3 & 25.2 &  34.25 &	11.00 &	\textbf{15.25} \\
        \rowcolor{gray!15} EVA01 ViT-g-14~\citep{sun2023evaclipimprovedtrainingtechniques}& 224$^2$ & 1136.4 & 78.5 & 6.7 & 26.7 & 40.0 & 6.7 & 13.3 & \textbf{66.7} & 13.3 & 13.3 & 20.0 & 23.0 & 27.25&	9.25 &	11.25 \\
        EVA02 ViT-bigE-14+~\citep{sun2023evaclipimprovedtrainingtechniques}& 224$^2$ & 5044.9 & 82.0  & 13.3 & 20.0 & 66.7 & 26.7 & 26.7 & \textbf{66.7} & 26.7 & 20.0 & 33.3 & 33.3 & 32.00 & 10.50 &	13.50 \\
        \midrule
        \rowcolor{gray!15} SD-v2.1-base~\cite{rombach2022high} & 512$^2$ &  865.9 & - & \textbf{31.1} & \textbf{33.3} & 35.6 & 20.0 & \textbf{33.3} & 46.7 & 33.3 & \textbf{33.3} & 44.4 & 34.6 & 31.92 & \textbf{14.17} & 10.50 \\
        \bottomrule
    \end{tabular}
    }
    \label{tab:mmvp_winoground_vlm}
    \vspace{-0.1in}
\end{table*}

\section{How Does Text Guidance Interact with Diffusion Representations?}
% This section investigates how text guidance modulates diffusion representations. We begin by analyzing diffusion cross-attention maps to assess the quality of vision-language alignment, then examine how this information propagates to spatial features, and finally evaluate its impact on alignment with the downstream language model.

\subsection{Cross-Attention Maps Capture Text-Aligned Visual Semantics}
\label{subsec:crossattn}

We first visualize internal cross-attention maps to assess how effectively diffusion models capture image-text correspondences. Using two COCO images~\cite{DBLP:journals/corr/ChenFLVGDZ15} and their captions, we compute averaged cross-attention maps at a $16 \times 16$ resolution. To study the effect of noise, we extract maps at early (low) and late (high) timesteps. As shown in Fig.~\ref{fig:cross_attn_maps_vis}, the model accurately grounds text in visual regions—for example, “racquets” highlights the correct object. We also find that attention maps across timesteps are complementary: higher timesteps emphasize broader contexts (e.g., “court,”), while lower timesteps focus on finer details. This aligns with diffusion dynamics, where early steps model global structure and later steps refine content.

To quantify alignment, we adopt image-text matching as a proxy task, following~\cite{he2023discffusion}. We compute scalar alignment scores by applying LogSumExp pooling~\cite{blanchard2019accuratecomputationlogsumexpsoftmax} over summed cross-attention maps across layers and timesteps. Based on our prior analysis, we aggregate maps from five representative timesteps ($t \in {189, 389, 589, 789, 989}$) and average results across three trials. We evaluate our method on MMVP-VLM~\cite{tong2024eyes} and Winoground~\cite{thrush_and_ross2022winoground}, two benchmarks that require fine-grained understanding of spatial relations, attributes, and compositional reasoning.

As shown in Table~\ref{tab:mmvp_winoground_vlm}, diffusion cross-attention maps outperform CLIP-based models across all benchmarks with a 5\% improvement over CLIP-ViT-H on MMVP-VLM and a 2.1\% improvement on Winoground. On MMVP-VLM, diffusion excels at tasks involving orientation (+11\%), identification of specific features (+20\%) and viewpoint (+17\%). This demonstrates that cross-attention capture robust vision-language alignments.   However, diffusion does not yet surpass specialized models like DFN-CLIP~\cite{fang2023datafilteringnetworks} and EVA-CLIP~\cite{sun2023evaclipimprovedtrainingtechniques}. Additional results, including per-timestep performance and ablations on timestep selection and noise sampling, are included in the Appendix.

\begin{figure}
    \centering
    \includegraphics[width=1\linewidth]{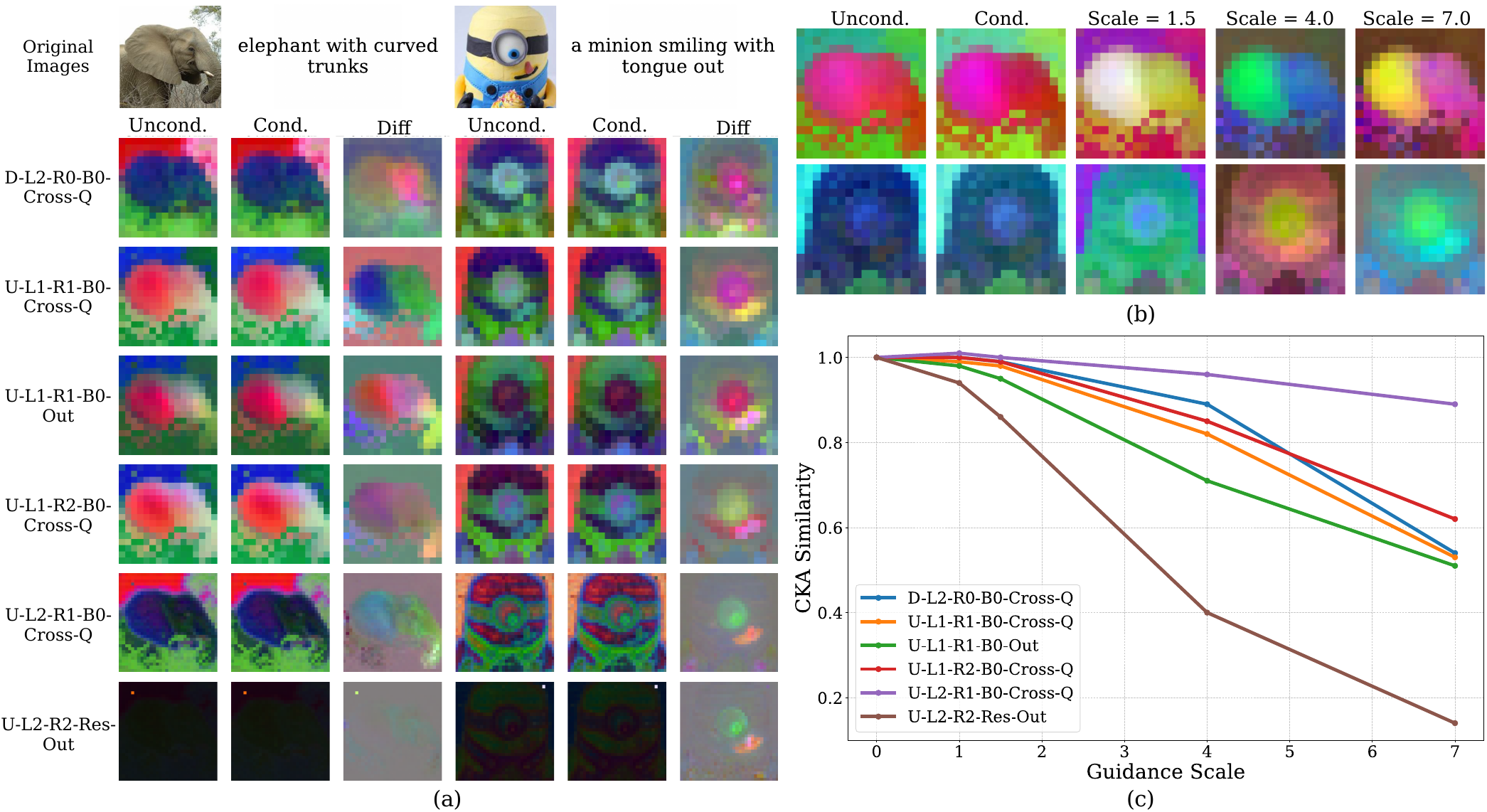}
    % \caption{\textbf{Visualizing Text Conditioned Diffusion Features.} (a) We sample a few images from the MMVP-VLM dataset and visualize PCA maps of spatial features extracted from different intermediate layers under both unconditional and text-conditioned settings (t=50). While the overall structure of the features remains similar after adding text, we find that the difference of these features can highlight specific regions influenced by the text. (b) We show the effect of amplifying text guidance—these affected regions become more prominent, while irrelevant areas are increasingly suppressed. (c) Finally, we measure CKA between unconditioned and progressively text-conditioned features across blocks and observe that spatial features extracted from \texttt{out} layers experience greater text modulation compared to \texttt{cross-q} features.}
    \caption{\textbf{Visualizing Text Conditioned Diffusion Features.} (a) A few images are sampled from the MMVP-VLM dataset and visualize PCA maps of spatial features extracted from different intermediate layers under both unconditional and text-conditioned settings (t = 50). While the overall structure of the features remains similar after adding text, we find that the difference of these features ($\Delta_{\text{cond}}$) can highlight specific regions influenced by the text. (b) We show amplifying text guidance highlights relevant object and part regions. (c) Finally, we measure CKA similarity between unconditioned and progressively text-conditioned features across blocks and observe that spatial features extracted from \texttt{out} layers experience greater text modulation compared to \texttt{cross-q} features.}
    
    \label{fig:pca_analysis_diff_features_cond}
\end{figure}

\subsection{Text Conditioning Modulates Spatial Feature Representations}
\label{subsec:text_spatial}

% Given the strong image-text associations observed in the cross-attention maps, we next investigate whether this information influences the underlying spatial feature representations. Specifically, we examine how text conditioning modulates the block-level features. 
Given the strong image-text associations observed in the cross-attention maps, we next investigate how text conditioning modulates the block-level features. We select a few images from the MMVP-VLM dataset and inspect PCA maps for both unconditional and conditional features across different blocks, with the timestep set to 50. We also visualize the PCA maps of the difference between the two features. The results are shown in Fig~\ref{fig:pca_analysis_diff_features_cond}(a). We make a few key observations. First, we find that there are minimal structural changes between the unconditional and conditional features, as the PCA maps of both remain relatively similar across all blocks for all samples. However, upon inspecting the difference, we find that the change between conditional and unconditional features can highlight key attributes of the image, for instance, the elephant's ears and trunk are highlighted in the first image and the minion's tongue is localized in the second image. 

To understand this effect further, we visualize how the feature changes when we amplify text-guidance. This is done via the following equation 

\begin{equation}
    X_{\text{amplified}} = X_\text{uncond} + s (X_\text{cond} - X_{\text{uncond}})  
\end{equation}

\begin{figure}[tbp]
    \centering
    \begin{minipage}{0.4\linewidth}
        \centering
        \includegraphics[width=\linewidth]{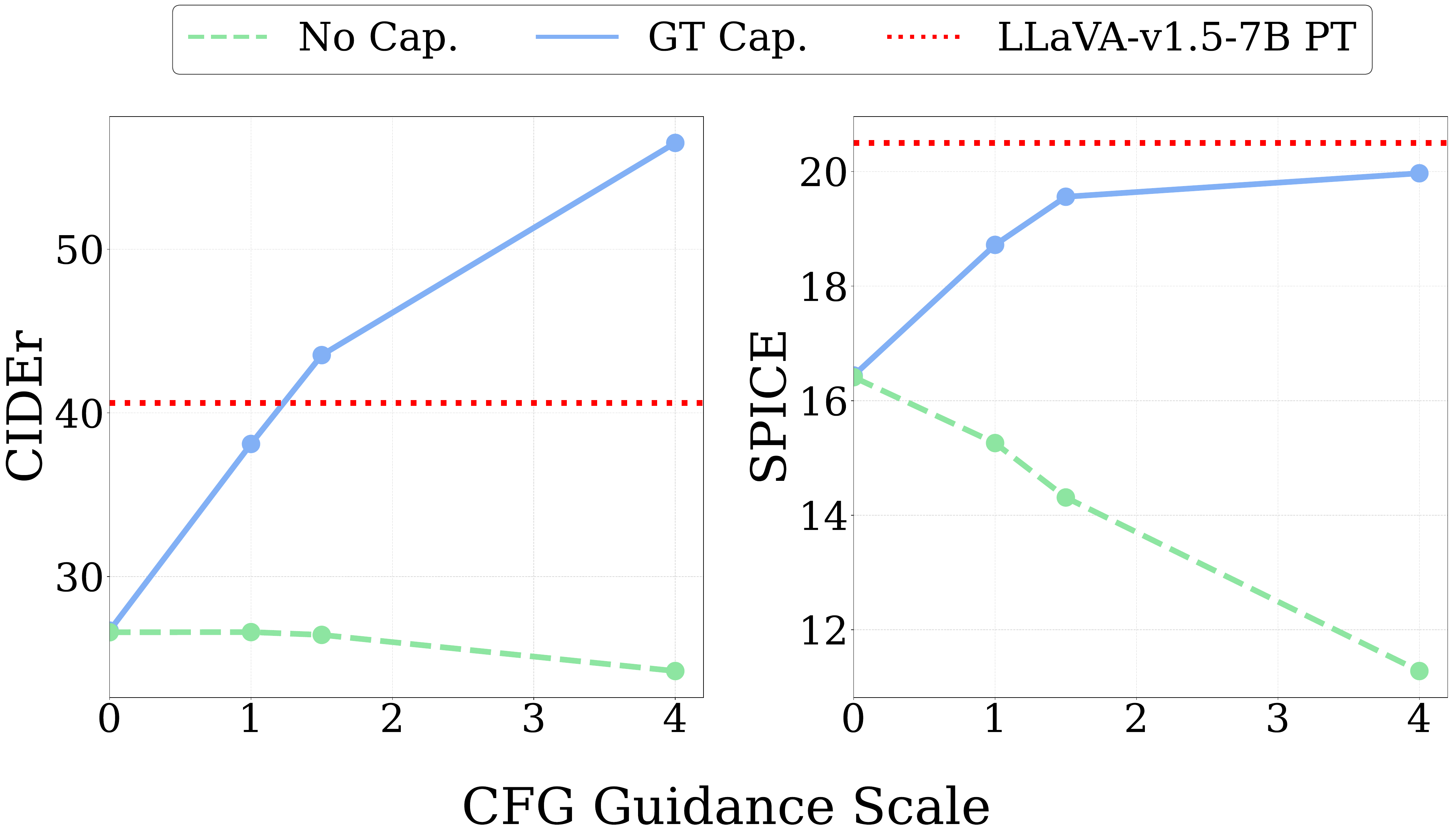}
    \end{minipage}
    \hfill
    \begin{minipage}{0.57\linewidth}
        \centering
        \includegraphics[width=\linewidth]{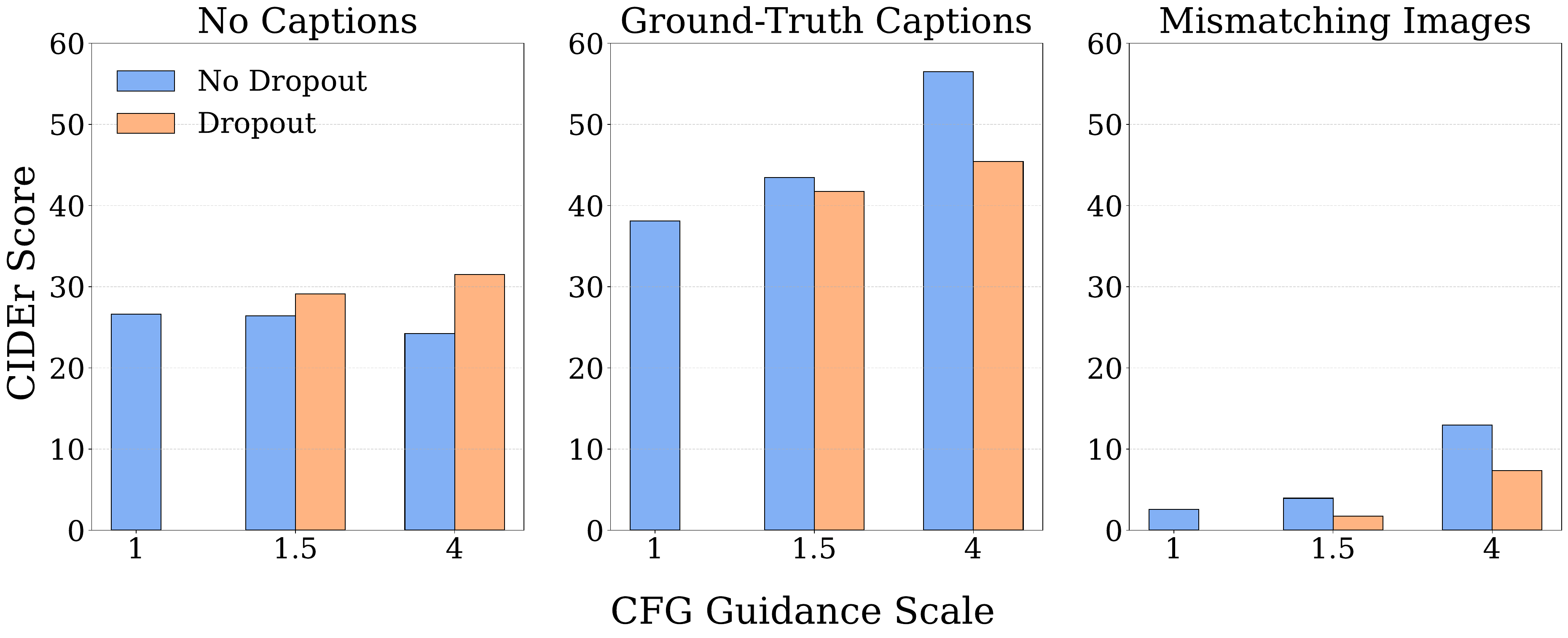}
    \end{minipage}
    
    % \caption{\textbf{COCO-Captions Performance} (left): We compare model performance on COCO-Captions across different pretraining text-guidance settings. Models trained with stronger ground-truth conditioning outperform the CLIP baseline when given ground-truth captions at inference but degrade significantly when no caption is provided. \textbf{Evidence of Leakage} (right): We examine whether increasing text-guidance scale causes prompt leakage into the LLM. Training with caption dropout improves robustness when no caption is provided, with minor performance drops when ground-truth captions are used. In a \textit{Mismatched} setting—where an unrelated image is paired with the caption—we find high captioning scores indicate that the diffusion features leak the input prompt to the LLM, enabling it to hallucinate content unrelated to the image. Applying dropout can mitigate this behavior.}
     \caption{\textbf{COCO-Captions Performance (Left)}: We compare model performance on COCO-Captions across different pretraining text-guidance settings. Models trained with stronger ground-truth conditioning outperform the CLIP baseline when given ground-truth captions at inference but degrade significantly when no caption is provided. \textbf{Evidence of Leakage (Right):} We examine whether increasing text-guidance scale causes prompt leakage into the LLM. Training with caption dropout improves robustness when no caption is provided, with minor performance drops when ground-truth captions are used.}
    \label{fig:coco_captions_performance}
\end{figure}

where $s$ is the guidance-scale and controls the amplification. We visualize the change for \texttt{U-L1-R1-B0-Cross-Q} as we increase $s$ from 0 (purely unconditional) to 7 and show the results in Fig.~\ref{fig:pca_analysis_diff_features_cond}(b). We find that while lower $s$ settings do not modify the structure of the image, high $s$ values increase focus on different parts o   f the image such as the ears for the elephant or the mouth for the minion. 

We quantify the effect of this change by performing a CKA analysis~\cite{kornblith2019similarity} between  unconditional features and conditional features across various guidance scales using the COCO-captions test-set. Our intuition is that features that experience high text modulation will have lower CKA similarity with the unconditional representation as guidance-scale increases. Our results are shown in Fig.~\ref{fig:pca_analysis_diff_features_cond}(c). We observe that all blocks experience substantial text modulation where \texttt{res-out} and \texttt{out} features experience the greatest change while \texttt{cross-q} features show more moderate effects. 

\subsection{Amplified Text Guidance Enables Leakage}
\label{subsec:leakage}

% Observing the significant text modulation capabilities by adjusting the guidance-scale, in this section, 

Based on sensitivity to guidance-scale, we aim to empirically determine how increasing guidance changes model performance. Specifically, if increasing text-guidance can improve alignment between the diffusion features and the downstream LLM. For this analysis, we use image captioning as a proxy task and train only the projection layer on the LLaVA-558K dataset. Our intuition is that improved alignment between diffusion features and the LLM will translate to better captioning performance. 

All models are trained with only the  $16 \times 16$ resolution \texttt{U-L1-R1-B0-Cross-Q} feature at t = 50. We choose this layer as it exhibits good unconditional performance and also experiences reasonable text-modulation. We choose to provide the ground-truth caption as input during training as an oracle to show the potential for text-guidance. For evaluation, we use the COCO-Captions test-set and use CIDEr~\cite{vedantam2015cider} and SPICE~\cite{anderson2016spice} metrics. During inference, we measure performance when passing ground-truth captions (GT Cap.) and no captions (No Cap.). The results are shown on the left of Fig~\ref{fig:coco_captions_performance}. We observe that increasing guidance-scale during training results in improved performance when the ground-truth is provided, surpassing even CLIP. However, these models show worse performance when no caption is given. This inverse trend suggests the possibility that the LLM may actually be able to extract the input text-prompt from the diffusion features. 

We investigate this phenomenon further via a mismatched setting. Namely, during inference, for an unrelated image-text pair, if the LLM is able to reconstruct the caption, then it is evidence of leakage. The results of this are shown on the right in Fig~\ref{fig:coco_captions_performance} (Mismatching Images) and illustrate that this is the case as the model trained with $s = 4$ is able to achieve a CIDEr score of 12.97. As a mitigation, we implement dropout during training where no caption is passed to the diffusion model randomly during pre-training. We observe that this improves robustness as captioning performance decreases on the Mismatching setting and performance on the No-Captions setting remains consistent across guidance-scales. This shows that dropout training helps the model learn a better balance between extracting image features vs. simply learning to decode the text information present in diffusion features. This also aligns with how the diffusion models are originally trained with classifier-free guidance, where the conditional prompt is occasionally masked for better generative performance. 

\section{Can We Extract Task-Aware Features for Question-Answering?}
\label{sec:vqa}

\begin{figure}[tbp]
\centering
\begin{minipage}[h]{0.48\textwidth}
    \centering
    \includegraphics[width=0.98\textwidth]{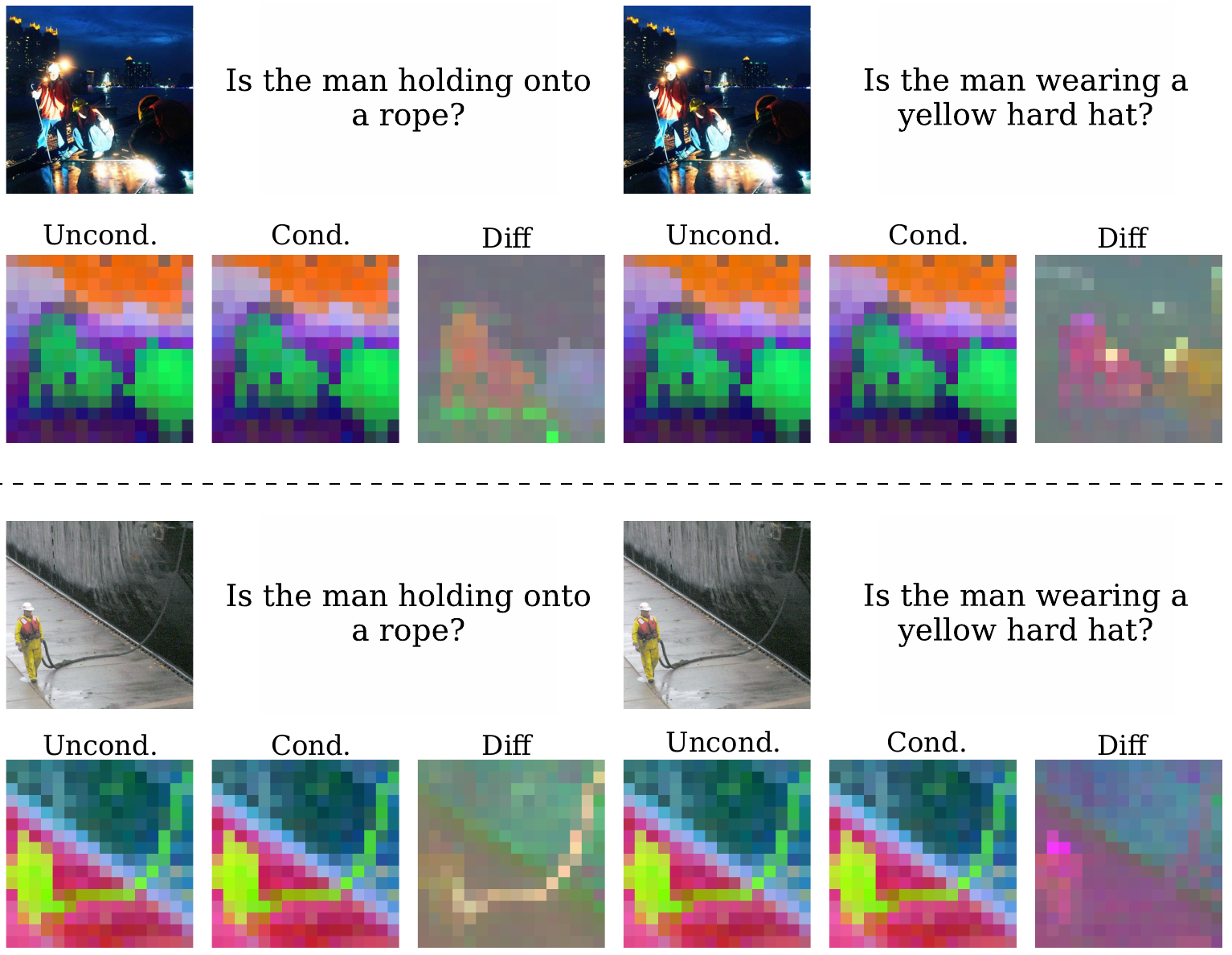}
    \caption{\textbf{Visualizing Question-Conditioned Features.} We sample an image with two questions from NaturalBench and visualize spatial features conditioned on questions via PCA. We see that the difference between conditional and unconditional features finds the relevant regions.}
\label{fig:pca_analysis_diff_features_question}
\end{minipage}
\hfill
\begin{minipage}[h]{0.49\textwidth}
    \centering
    \scriptsize % Using smaller font size (scriptsize instead of footnotesize)
    \begin{tabular}{@{}l@{\hspace{1mm}}c@{\hspace{1mm}}c@{\hspace{1mm}}c@{\hspace{1mm}}c@{\hspace{1mm}}c@{\hspace{1mm}}c@{\hspace{1mm}}c@{}}
    \toprule
    \multirow{2}{*}{\textbf{Config}} & \textbf{LLaVA-B} & \textbf{MMVP} & \textbf{GQA} &  \multicolumn{3}{c}{\textbf{Natural-Bench}} \\
    \cmidrule(r){2-2} \cmidrule(r){3-3} \cmidrule(r){4-4} \cmidrule(){5-7}
    & All & Acc & Acc & Q-Acc & I-Acc & G-Acc \\
    \midrule
    LLaVA-v1.5-7B & 66.5 & 24.7 & 62.7 & 37.70 & 43.80 & 14.32 \\
    \midrule
    \multicolumn{6}{@{}l}{\textit{SD \& CLIP Concat.}} \\
    \midrule
     Scale = 0 (Uncond.) & 64.2 & 26.7 & 63.1 & 39.03 & 44.61 &	13.84 \\
     Scale = 4 & 61	& 27.3 &  \textbf{63.2} & 37.89 &	43.50 & 12.84 \\
    \midrule
    \multicolumn{6}{@{}l}{\textit{SD \& CLIP Cross-Attn.}} \\
    \midrule
    Scale = 0 (Uncond.) & \textbf{66.8} & 26.9 & 62.7 & 40.55 & 46.13 & 14.79 \\
    Scale = 4 & 63.7 & \textbf{30.7} & 62.5 & \textbf{40.71} & \textbf{46.21} & \textbf{15.26} \\
    \bottomrule
    \end{tabular}
    \captionof{table}{\textbf{Performance using CLIP and Conditional Diffusion Features.} We evaluate model performance when fusing CLIP and diffusion features and demonstrate improved performance when using conditional features at $s = 4$ on the MMVP and NaturalBench benchmarks.}
    % \vspace{-36pt}
    \vspace{15pt}
    \label{tab:clip_sd_performance}
\end{minipage}
\end{figure}

\subsection{Conditioning with Questions Focuses on Relevant Regions}
\label{subsec:question_cond}

We now turn our focus to using questions as input to the diffusion model to extract question-aware features. While our previous analysis showed that amplifying text guidance with the caption increased focus on caption-related regions, we investigate whether this holds for question inputs. We sample a pair of images from the NaturalBench benchmark with corresponding questions~\cite{naturalbench}. We perform PCA analysis of the unconditional and conditional features and their difference. As shown in Fig.~\ref{fig:pca_analysis_diff_features_question}, we observe that the difference between conditional and unconditional features highlight regions that match with the question. On the top row, we see that when asking about the rope, the pipe regions are highlighted, while when asking about whether the man is wearing a yellow hat, greater focus is placed on the top of each man's head. We observe that these trends hold for the bottom image as well. 

\subsection{Fusing Conditional Diffusion and CLIP Improves Multimodal Understanding}
\label{subsec:clip_sd}

Based on our analyses, we propose leveraging CLIP and conditional diffusion features for improved multimodal understanding. We experiment with two fusion strategies, namely feature concatenation and cross-attention. For cross-attention, we use the CLIP features as the queries and the diffusion features as the keys and values. To prevent the large language model from incorrectly latching onto the text, we only pass question prompts during the SFT stage of training and provide no text during the pre-training stage. We provide results with both unconditional and question-conditioned diffusion features at $s = 4$ as shown in Table~\ref{tab:clip_sd_performance}. We find that simple concatenation with unconditional diffusion features is able to improve on MMVP and GQA with a 2 point and 0.5 point increase in performance respectively. Increasing guidance to $s = 4$, we observe further improvement. With cross-attention fusion, we observe that performance on MMVP improves by +7 points and +0.9 points on NaturalBench compared to the original LLaVA-v1.5 baseline. 

\section{Discussion}
\label{sec:conclusion}

\noindent\textbf{Broader Impacts and Limitations.} 
%In this work, we utilize the Stable Diffusion v2.1 model, trained using a subset of the LAION-5B. 
%This dataset contains content depicting violence and sexually explicit material. and as such the dataset was filtered to remove these images prior to training. 
%However, there are still potential risks.
Our use of Stable Diffusion v2.1 is limited to feature extraction rather than image generation, so we expect the associated risks to be substantially reduced. 
%Additionally, our approach aims to reduce reliance on multiple feature extractors and thereby improve model efficiency.
LLM usage also has risks with respect to hallucinating content and we advocate for careful usage of these models. 
%In order to mitigate leakage effects, we chose to not pass text prompts during pre-training. 
%To better understand the effects of using text at this stage, we analyze the performance of models trained with amplified text-guidance across different guidance scales on NaturalBench. 
%The results as shown in Fig.~\ref{fig:natbench} demonstrate that while increasing guidance degrades performance relative to using unconditional diffusion features, there are non-overlapping questions which each model gets correct. 
While in the pipeline we study, text leaks can exacerbate the hallucination problem, we envision that an ideal ensembling strategy for leveraging each guidance scale's features could achieve better performance than CLIP and be explored as future work. 

\noindent\textbf{Conclusion.} 
In this work, we investigate the potential diffusion models have as task-aware feature extractors.
We show that diffusion features encode rich vision-centric information that enables improved performance on vision-centric benchmarks. 
We then explore how text-conditioning modulates the internal diffusion representations and showcase that its cross-attention maps capture fine-grained vision-text correspondences that propagate to block-level features. 
Finally, we present a simple approach to utilize the complementary information in CLIP and conditional diffusion features and showcase improved multimodal performance on general-purpose and vision-centric benchmarks.

\noindent\textbf{Acknowledgements.}
This work was partially supported by NSF CAREER Award (\#2238769). The authors would like to thank our colleagues Anubhav Gupta, Soumik Mukhopadhyay, and Pulkit Kumar for their valuable conversations and feedback. The authors acknowledge UMD’s supercomputing resources made available for conducting this research. The U.S. Government is authorized to reproduce and distribute reprints for Governmental purposes notwithstanding any copyright annotation thereon. The views and conclusions contained herein are those of the authors and should not be interpreted as necessarily representing the official policies or endorsements, either expressed or implied, of NSF or the U.S. Government.

\clearpage
{\small
\bibliographystyle{ieeetr}
\bibliography{main}
}

\clearpage
\appendix
% --- PDF will be split by an editor (e.g. macOS preview), so need to restart from page 1
\setcounter{page}{1}
\setcounter{section}{0}

%\twocolumn[
{
\centering
\Large
\textbf{Towards Multimodal Understanding via
Stable Diffusion as a Task-Aware Feature Extractor} \\ 
     {\small Supplementary Material} \\
\vspace{2.0em}
}

\section{Experimental Settings}

\subsection{Block Configurations}
We first describe our block-selection settings in more detail. We extract features at both the encoder (\texttt{down-stage}) and the decoder (\texttt{up-stage}). Additionally, we inspect mostly pixel-wise query features used in the text-conditioned cross-attention layers. We inspect these features across blocks and resolutions. Lastly, we examine a select set of output features (denoted as \texttt{out}). We adopt a structured naming convention to identify specific feature blocks within the diffusion model. Each name follows the format: \texttt{Stage-Level-Repeat-Block-FeatureType}. The stage is either D or U, followed by the resolution level (L\#), the residual block index level (R\#), the transformer block index (B\#), and finally the type of feature extracted (Cross-Q, Out, or Res-Out). This notation allows for precise referencing of intermediate features across the model hierarchy. The full configurations are shown in Table~\ref{tab:block_configs}. 

\begin{table*}[h]
\begin{minipage}{0.65\linewidth}
\caption{\textbf{Block Configurations.} We describe all block configurations used in our analysis and experiments.}
\label{tab:block_configs}
\resizebox{\textwidth}{!}{%{\scriptsize 
\setlength{\tabcolsep}{5pt}
\renewcommand{\arraystretch}{1.2}
\begin{tabular}{@{}l|c|c|>{\raggedleft\arraybackslash}p{6.5cm}@{}}
    \toprule
    Feature Config. & Res & Dim & Description \\
    \midrule
    D-L2-R0-B0-Cross-Q & $16 \times 16$ & 1280 & Down-Stage, Level 2, 1st ResBlock, Cross-Attention Query \\
    U-L1-R1-B0-Cross-Q & $16 \times 16$ & 1280 & Up-Stage, Level 1, 2nd ResBlock, Cross-Attention Query \\
    U-L1-R1-B0-Out     & $16 \times 16$ & 1280 & Up-Stage, Level 1, 2nd ResBlock, Attention Output \\
    U-L1-R2-B0-Cross-Q & $16 \times 16$ & 1280 & Up-Stage, Level 1, 3rd ResBlock, Cross-Attention Query \\
    U-L2-R1-B0-Cross-Q & $32 \times 32$ & 640  & Up-Stage, Level 2, 1st ResBlock, Cross-Attention Query \\
    U-L2-R2-Res-Out    & $32 \times 32$ & 640  & Up-Stage, Level 2, 3rd ResBlock, Output \\
    \bottomrule
\end{tabular}
}
\end{minipage}%
\hfill%
\begin{minipage}{0.3\linewidth}
\caption{\textbf{Training Hyperparameters.} Format from ~\cite{tong2024eyes}}.
\label{tab:llava-hyperparams}
\resizebox{\textwidth}{!}{%
\setlength{\tabcolsep}{4pt}
\renewcommand{\arraystretch}{1.2}
\begin{tabular}{l|cc}
    \toprule
    \textbf{Hyperparameter} & \multicolumn{2}{c}{\textbf{LLaVA-1.5}} \\
    & Stage 1 & Stage 2 \\
    \midrule
    batch size         & 256  & 128 \\
    lr                 & 2e-3 & 2e-5 \\
    lr schedule decay  & cosine & cosine \\
    lr warmup ratio    & 0.03 & 0.03 \\
    weight decay       & 0    & 0 \\
    epoch              & 1    & 1 \\
    optimizer          & \multicolumn{2}{c}{AdamW~\cite{loshchilov2019decoupledweightdecayregularization}} \\
    DeepSpeed stage    & 2    & 2 \\
    \bottomrule
\end{tabular}
}
\end{minipage}
\end{table*}

\subsection{LLaVA Training Settings}

For visual feature processing, we fix the number of tokens to 256 and resize all features to $16 \times 16$ regardless of their original resolution. Our experimental settings follow LLaVA~\cite{liu2023llava, liu2023improvedllava}. The full hyperparameters are described in Table~\ref{tab:llava-hyperparams}. Our training protocol remains mostly unchanged, except we use DeepSpeed stage 2 for both stages of training. We train using 4 NVIDIA H100s. For our LLM, we use the Vicuna-7B~\cite{chiang2023vicuna} model. 

\section{Full Captioning Results}

We tabulate all COCO-Caption results displayed in Fig.~\ref{fig:coco_captions_performance} and present them in Table~\ref{tab:coco_captions}. 

\begin{table*}[ht]
  \setlength{\cmidrulewidth}{0.02em}
  \renewcommand{\tabcolsep}{10pt}  % column separation
  \renewcommand{\arraystretch}{1.2}% row height
  \centering
  \caption{Comparison of models on the COCO‑Captions benchmark. 
           SD2.1 uses $512{\times}512$ images; CLIP uses $336{\times}336$ images.}
  \vspace{-0.7\baselineskip}      % tighten vertical space (optional)

  % Make the whole table as wide as the text block
  \resizebox{\textwidth}{!}{%
  \begin{tabular}{
      >{\hspace{-\tabcolsep}}l      % remove the extra padding at the left edge
      ccc                            % 3 columns under “Model Details”
      cccc<{\hspace{-\tabcolsep}}    % 4 metric columns, then remove right padding
  }
    \toprule
    \multicolumn{3}{c}{\textbf{Model Details}} &
    \multicolumn{4}{c}{\textbf{COCO‑Captions Benchmark}} \\
    \cmidrule(r){1-3}\cmidrule(l){4-7}
    Model & Train Mode & Val Mode &
    ROUGE‑L & CIDEr & B@4 & SPICE \\
    \midrule
    CLIP-ViT-L14-336 & No Captions & No Captions & 
    40.75 &	40.60 & 15.02	& 20.50   \\
    \midrule
    \rowcolor{gray!15}
    Stable‑Diffusion‑2.1‑base (PT) & No Captions & No Captions &
      36.78	& 26.59	& 11.77	& 16.41 \\
    Stable‑Diffusion‑2.1‑base (PT) & GT Captions & No Captions &
      34.27	& 26.60 &	11.13 & 15.26 \\
    \rowcolor{gray!15}
    Stable‑Diffusion‑2.1‑base (PT) & GT Captions & GT Captions &
      39.31	& 38.10	& 14.23	& 18.72 \\
    \midrule
    Stable‑Diffusion‑2.1‑base (PT) &  GT ($s = 1.5$) & No Captions &
      33.43	& 26.43	& 10.53	& 14.31 \\
    \rowcolor{gray!15}
    Stable‑Diffusion‑2.1‑base (PT) &  GT ($s = 1.5$) & GT‑Captions &
     40.70	& 43.53	& 15.46	 & 19.56 \\
    \midrule
    Stable‑Diffusion‑2.1‑base (PT) &  GT ($s = 1.5$ w/ Dropout) & No Captions &
      36.13	& 29.10	& 11.99	& 16.10 \\
    \rowcolor{gray!15}
    Stable‑Diffusion‑2.1‑base (PT) &  GT ($s = 1.5$ w/ Dropout) & GT‑Captions &
     40.77 & 41.72	& 15.22	& 19.46\\
    \midrule
    Stable‑Diffusion‑2.1‑base (PT) &  GT ($s = 4$) & No Captions &
      31.08	& 24.22	& 9.40 &	11.28 \\
    \rowcolor{gray!15}
    Stable‑Diffusion‑2.1‑base (PT) & GT ($s = 4$) & GT‑Captions &
      43.74	& 56.50 & 18.10 & 19.97 \\ 
    \midrule
    Stable‑Diffusion‑2.1‑base (PT) & GT ($s = 4$) w/ Dropout & No‑Captions &
      37.52 &	31..48	& 12.49	& 16.10 \\ 
    \rowcolor{gray!15}
    Stable‑Diffusion‑2.1‑base (PT) & GT ($s = 4$) w/ Dropout & GT‑Captions &
      42.44	& 45.40	& 16.19	& 20.04 \\ 
    \bottomrule
  \end{tabular}}%
  \label{tab:coco_captions}
  \vspace{-0.8\baselineskip}       % tighten space after the table (optional)
\end{table*}

\section{Further Diffusion Feature Analysis}

% \begin{figure}[h]
%     \centering
%     \framebox(120,180){}
%     \framebox(120,180){}
% \end{figure} 

\subsection{Visualizations}

We provide more visualizations of unconditional diffusion features via the NaturalBench benchmark~\cite{naturalbench} as it consists of more complex image-pairs compared to MMVP~\cite{tong2024eyes}. We observe similar trends as our analysis in Sec~\ref{subsec:uncond_analysis}. We observe more instances of shared objects represented similarly across each pair of images (as evidenced by the similar colors of the PCA maps). For instance, in pair (d), diffusion features across multiple blocks capture shared semantics of the motorcycle, its wheels, and the driver. Other examples include the shared encoding of the face masks in pair (e) for U-L1-R1-B0-Cross-Q and U-L1-R1-B0-Out. Interestingly, we see that this behavior is not as present in features extracted at the \texttt{down-stage}. We hypothesize that this could be a result of the diffusion noises present during the encoding stage. Lastly, we note more examples of ``register'' token behavior for \texttt{vit-out} and \texttt{res-out} features. This can be seen for all samples for the \texttt{res-out} features and pairs (a-d) for the \texttt{vit-out} features.  

\begin{figure}[htbp]
    \centering
    \includegraphics[width=1\linewidth]{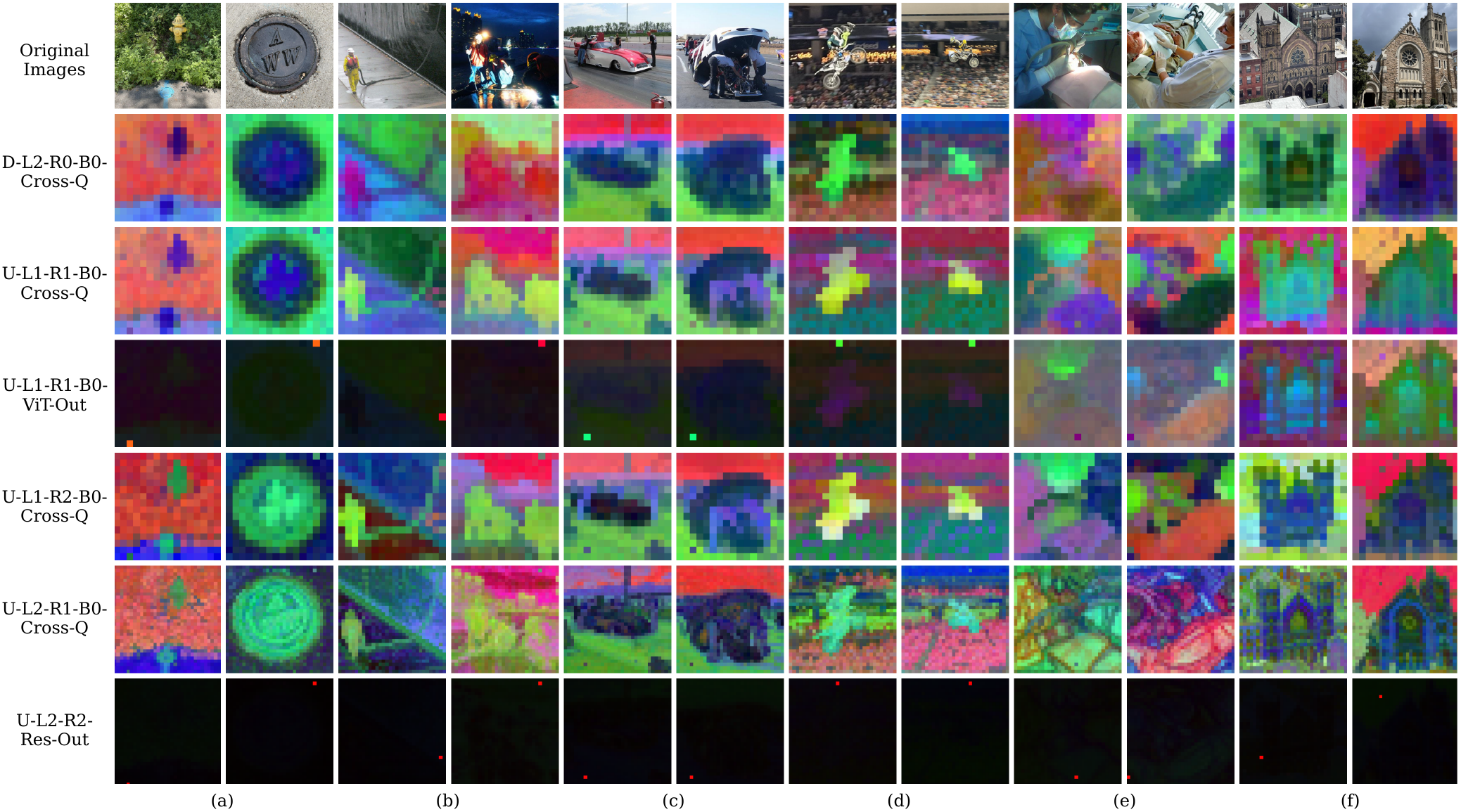}
    \caption{\textbf{More Diffusion Feature Visualizations.} We sample six pairs of images from the NaturalBench benchmark~\cite{naturalbench} and view the joint PCA maps across different blocks and layers. Please zoom in to see more details.}
    \label{fig:pca_analysis_diff_features_uncond_nb}
\end{figure}

\subsection{CKA Analysis}

\begin{figure}[ht]
    \centering
    \includegraphics[width=1\linewidth]{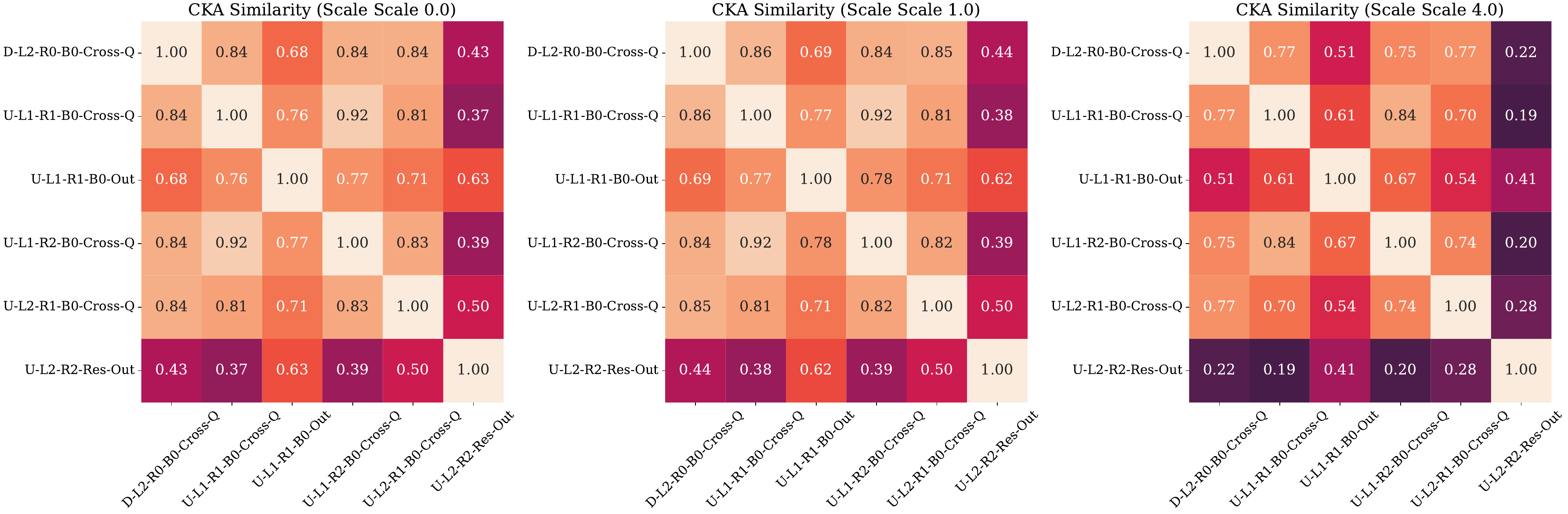}
    \caption{\textbf{CKA Block-Wise Similarity.} We compute block-wise CKA similarity using the COCO-Captions test set (5000 images) across various guidance scales ($s = $ 0, 1, 4)}
    \label{fig:blockwise_cka}
\end{figure}

Here, we aim to understand the relationship between representations across different blocks. Specifically, we use CKA~\cite{kornblith2019similarity} to measure the similarity of different representations. We use the COCO-Captions test set, which consists of 5000 images, for our analysis. Furthermore, we explore how increasing text guidance may impact these block-wise relationships by computing CKA at various guidance scales, namely $s = 0,1,4$. As shown in Fig.~\ref{fig:blockwise_cka}, for $s = 0$ (purely unconditional features) \texttt{cross-q} features exhibit more similar representations compared to \texttt{out} features. When conditioning on text, we observe minimal changes in feature similarities. However, increasing text guidance ($s = 4$), we see that pair-wise CKA similarity decreases for all pairs of blocks, with a greater decrease for \texttt{out} blocks. This aligns with the trends we observed in Fig.~\ref{fig:pca_analysis_diff_features_cond} where \texttt{out} features exhibited greater changes in their representations due to higher text guidance. 

\section{Further Cross-Attention Map Analysis}

\subsection{Quantifying Effect of Timesteps}

\begin{table*}[htbp]
\setlength{\cmidrulewidth}{0.01em}
\renewcommand{\tabcolsep}{5pt} % Adjusted spacing
\renewcommand{\arraystretch}{1.2} % Slightly increased row height
\centering
\caption{Comparison of SD2.1 model across varying timesteps for MMVP-VLM Benchmark, using $512 \times 512$ images. For `Ensemble' we use timesteps $t \in \{189, 389, 589, 789, 989\}$, and average results across 3 trials.}
\vspace{-0.1in}
\small
\resizebox{\linewidth}{!}{
\begin{tabular}{>{\kern-\tabcolsep}lccccccccccc<{\kern-\tabcolsep}}
\toprule
\multicolumn{2}{c}{{Model Details}} & \multicolumn{10}{c<{\kern-\tabcolsep}}{{MMVP-Val Benchmark}} \\ 
\cmidrule{1-2} \cmidrule(l){3-12}
Model & Timesteps & \faCompass & \faSearch & \faSync & \faSortNumericUp & \faMapPin & \faPalette & \faCogs & \faFont & \faCamera & Avg \\ 
%Model & Timesteps & Noises & \faCompass & Feat. Presence & State/Cond. & Quality/Count & Pos./Rel. & Col./Appear. & Struct. & Texts & View. & Avg \\ 
\midrule

\rowcolor{gray!15} SD-v2.1 & 89 & 0.00 	&13.33	&20.00	&13.33	&\textbf{40.00}&33.33	&26.67	&26.67	&\textbf{46.67}&24.44
\\ % Row 14

SD-v2.1 & 189 & 20.00&13.33&26.67&6.67&26.67&20.00&20.00&33.33&20.00&20.74
\\ % Row 15

\rowcolor{gray!15} SD-v2.1 & 289 & 33.33&26.67&26.67&26.67&33.33&40.00&\textbf{40.00}&33.33&13.33&30.37
\\ % Row 16

SD-v2.1 & 389 & 33.33&26.67&33.33&20.00&13.33&26.67&\textbf{40.00}&20.00&33.33&27.41
\\ % Row 17

\rowcolor{gray!15} SD-v2.1 & 489 & 20.00&20.00&40.00&20.00&20.00&46.67&\textbf{40.00}&13.33&13.33&25.93
\\ % Row 18

SD-v2.1 & 589 & 20.00&\textbf{33.33}&\textbf{53.33}&26.67&33.33&33.33&20.00&33.33&20.00&30.37
\\ % Row 19

\rowcolor{gray!15} SD-v2.1 & 689 & 13.33&20.00&13.33&13.33&33.33&40.00&26.67&13.33&\textbf{46.67}&24.44
\\ % Row 20

SD-v2.1 & 789 & 26.67&13.33&33.33&13.33&\textbf{40.00}&46.67&\textbf{40.00}&\textbf{40.00}&13.33&29.63
\\ % Row 21

\rowcolor{gray!15} SD-v2.1 & 889 & 13.33&\textbf{33.33}&33.33&\textbf{46.67}&\textbf{40.00}&60.00&33.33&26.67&26.67&\textbf{34.81}
\\ % Row 22

SD-v2.1 & 989 & \textbf{46.67}&0.00&26.67&13.33&\textbf{40.00}&\textbf{66.67}&20.00&20.00&33.33&29.63
\\ % Row 23

\midrule
%\rowcolor{gray!15} SD-v2.1 & Ensemble & %46.67&33.33&40.00&13.33&13.33&40.00&33.33&13.33&46.67&31.11
%\\ % Row 24

%SD-v2.1 & Ensemble & 5 & 
%40.00&26.67&46.67&20.00&33.33&40.00&26.67&33.33&53.33&35.56
%\\ % Row 25

%\rowcolor{gray!15} SD-v2.1 & Ensemble & 5 & %26.67&33.33&13.33&26.67&46.67&60.00&33.33&46.67&40.00&36.30
%\\ % Row 26

%SD-v2.1 & Ensemble & 5 & 
%26.67&40.00&46.67&13.33&20.00&40.00&40.00&20.00&40.00&31.85
%\\ % Row 27

\rowcolor{gray!15} SD-v2.1 & Ensemble & 31.1 ± 7.70 & \textbf{33.3} ± 6.67 & 35.6 ± 19.25 & 20.0 ± 6.67 & 33.3 ± 13.33 & 46.7 ± 11.55 & 33.3 ± 6.67 & 33.3 ± 13.33 & 44.4 ± 7.70 & 34.6 ± 2.38
\\ % Row 28

\bottomrule
\end{tabular}
}
\vspace{-0.15in}
\label{tab:mmvp_time_ablation}
\end{table*}

We examine how different time-steps impact performance for MMVP-VLM in Table~\ref{tab:mmvp_time_ablation} and observe that features extracted from earlier timesteps yield better performance on fine-grained patterns, such as the presence of specific concepts, but worse performance for color and appearance. We do the same for the Winoground benchmark, as shown in Table~\ref{tab:winoground_time_ablation}. We find that earlier timesteps ($t \in \{89, 189, 289\}$) have a significantly lower ability to select the correct image given a caption (image-score) compared to later timesteps with a 14pt gap between features taken from t = 89 and
features from t = 989. Ensembling features across multiple timesteps helps balance out performance and we observe this strategy achieves the highest text-score (ability for model to select the correct caption given an image) with relatively good performance on image- and group-score as well.

\begin{table*}[htbp]
\setlength{\cmidrulewidth}{0.01em}
\renewcommand{\tabcolsep}{5pt} % Adjusted spacing
\renewcommand{\arraystretch}{1.2} % Slightly increased row height
\centering
\caption{Comparison of SD2.1 model across varying timesteps for Winoground Benchmark, using $512 \times 512$ images. For `Ensemble' we use timesteps $t \in \{189, 389, 589, 789, 989\}$, and average results across 3 trials with 5 noise-steps.}
\vspace{-0.1in}
\small
\resizebox{\linewidth}{!}{
\begin{tabular}{>{\kern-\tabcolsep}lcccccc<{\kern-\tabcolsep}}
\toprule
\multicolumn{2}{c}{{Model Details}} & \multicolumn{3}{c<{\kern-\tabcolsep}}{{Winoground Benchmark}} \\ 
\cmidrule{1-2} \cmidrule(l){3-6}
Model & Timesteps & Text & Image & Group \\  
%Model & Timesteps & Noises & \faCompass & Feat. Presence & State/Cond. & Quality/Count & Pos./Rel. & Col./Appear. & Struct. & Texts & View. & Avg \\ 
\midrule
\rowcolor{gray!15} Stable-Diffusion-v2.1-base (512) & 89 & 29.25 & 5.75 &	3.00
\\ % Row 13

Stable-Diffusion-v2.1-base (512) & 189 & 23.75 & 8.25 &	4.25
\\ % Row 14

\rowcolor{gray!15} Stable-Diffusion-v2.1-base (512) & 289 & 25.50 &	9.25 &	6.50
\\ % Row 15

Stable-Diffusion-v2.1-base (512) & 389 & 26.00  &	12.75 &	10.25
\\ % Row 16

\rowcolor{gray!15} Stable-Diffusion-v2.1-base (512) & 489 & 29.50 &	16.00 & 11.25
\\ % Row 17

Stable-Diffusion-v2.1-base (512) & 589 & 29.75 &	15.75 &	10.00
\\ % Row 18

\rowcolor{gray!15} Stable-Diffusion-v2.1-base (512) & 689 & 30.25 &	13.75 &	10.50
\\ % Row 19

Stable-Diffusion-v2.1-base (512) & 789 & 27.50 &   {17.00} &	12.75
\\ % Row 20

\rowcolor{gray!15} Stable-Diffusion-v2.1-base (512) & 889 & 27.50 &	12.50 & 9.00
\\ % Row 21

Stable-Diffusion-v2.1-base (512) & 989 & 28.00 &	\textbf{19.75} &	\textbf{14.25}
\\ % Row 22

% \rowcolor{gray!15} Stable-Diffusion-v2.1-base (512) & [189, 389, 589, 789, 989] - (5 noise) & 29.50 &	13.50 &	10.00
% \\ % Row 27

% Stable-Diffusion-v2.1-base (512) & [189, 389, 589, 789, 989] - (5 noise) & \underline{34.75} &	15.50 &	11.75
% \\ % Row 28

% \rowcolor{gray!15} Stable-Diffusion-v2.1-base (512) & [189, 389, 589, 789, 989] - (5 noise) & 31.50 &	13.50 &	9.75
% \\ % Row 29

\midrule

\rowcolor{gray!15} Stable-Diffusion-v2.1-base (512) & [189, 389, 589, 789, 989] & \textbf{31.92} ± 2.65 & 14.17 ± 1.15 & 10.50 ± 1.09
\\ % Row 30

\bottomrule
\end{tabular}
}
\vspace{-0.15in}
\label{tab:winoground_time_ablation}
\end{table*}

\subsection{More Visualizations}

\begin{figure}[htb]
    \centering
    \includegraphics[width=1\linewidth]{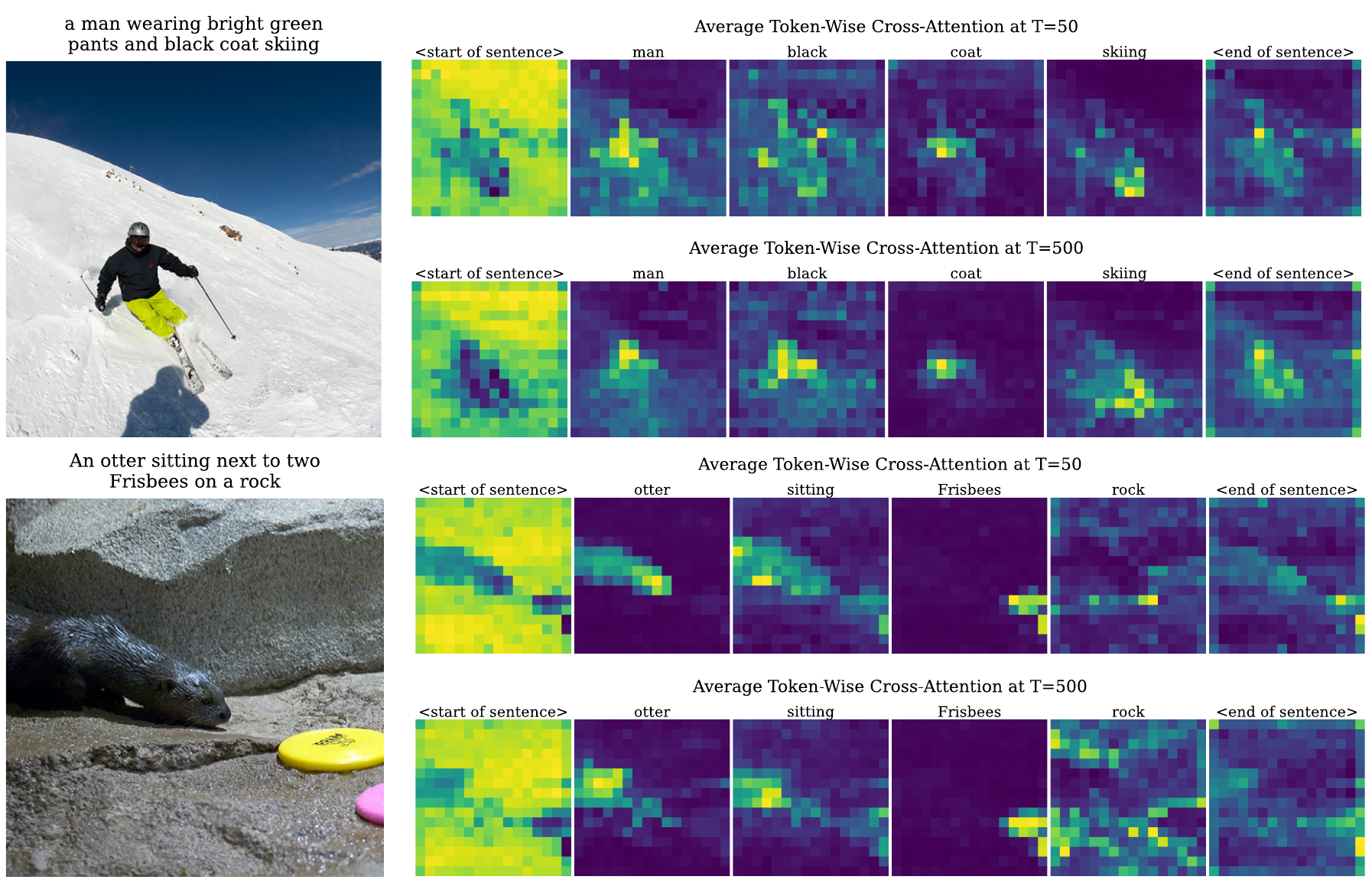}
    \caption{\textbf{More Cross-Attention Maps.} We display more examples of cross-attention maps from images in the COCO-Captions test set. For the image of the skier, we can see that object-attribute binding such as``black'' and ``coat'' are better aligned at later timesteps compared to earlier ones.}
    \label{fig:suppl_cross_attn_maps}
\end{figure}

\begin{figure}[h]
    \centering
    \includegraphics[width=1\linewidth]{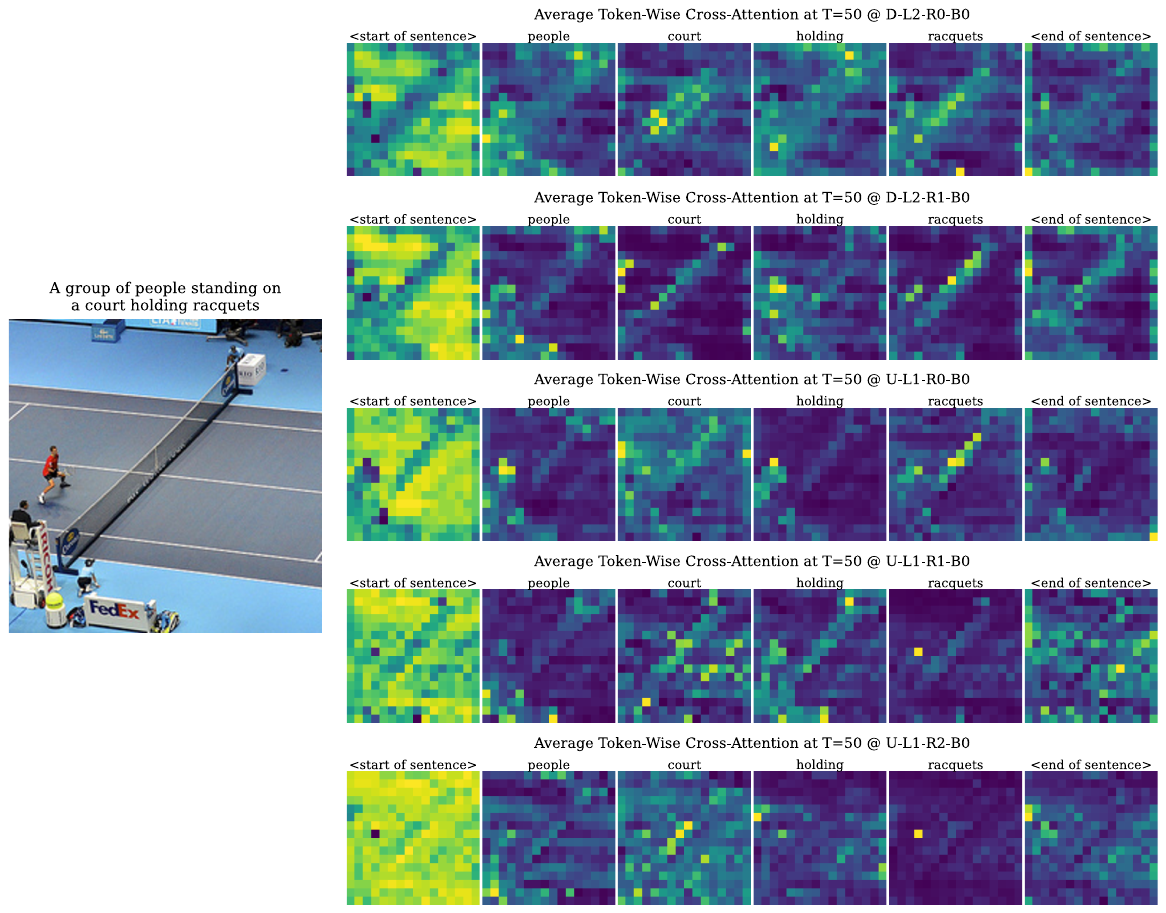 }
    \caption{\textbf{Cross-Attention Maps Across Layers.} We display cross-attention maps at timestep 50 across various layers. We see that cross-attention maps are not uniform and that maps at the \texttt{up-stage} encode more robust image-text alignment.}
    \label{fig:suppl_cross_attn_maps_layerwise}
\end{figure}

Here, we provide more visualizations of cross-attention maps at both early and later timesteps. We sample images from the COCO-Captions test set to generate these maps. We observe intriguing differences in how concepts are encoded across timesteps. For example, in the skier image, later timesteps more precisely capture object-attribute bindings (e.g., the word ``black'' better corresponds to the coat at t = 500, compared to t = 50). We also note that early timesteps may not always be more localized than higher timesteps as the map for ``coat'' is slightly more concentrated at t = 500 than t = 50. For the second image of the otter, we see that the ``Frisbees'' map precisely localizes the frisbees across both timesteps. We note similar trends concerning the higher timestep maps better-capturing backgrounds, as ``rock'' captures the background better at t = 500, while at t = 50, the corresponding map attends more on just the edge between the rocks. 

\subsection{Inspecting Cross-Attention across Layers}

While our earlier analysis averaged cross-attention maps across all layers at the $16 \times 16$ resolution, we now aim to understand how individual layers contribute to visual-linguistic representation. Specifically, we look at cross-attention maps at the following layers: \texttt{D-L2-R0-B0}, 
\texttt{D-L2-R1-B0},
\texttt{U-L1-R0-B0},  
\texttt{U-L1-R1-B0},  
\texttt{U-L1-R2-B0}. 
The maps are shown in Fig.~\ref{fig:suppl_cross_attn_maps_layerwise}. We observe that \texttt{down-stage} cross-attention maps vary considerably compared to \texttt{up-stage} maps. However, one set of maps is not necessarily better than the others. For instance, ``people'' captures the person on the court in the \texttt{U-L1-R0-B0} maps, while in the \texttt{D-L2-R1-B0} maps, ``people'' captures the people sitting on the left side of the court. Another example is the ``racquets'' maps which accurately localize the object in the \texttt{U-L1-R1-B0} and \texttt{U-L1-R2-B0} maps but focus on the court net in the rest of them. 

\section{More Examples of Question-Conditioning in Diffusion Features}

\begin{figure}[htbp]
    \centering
    \includegraphics[width=1\linewidth]{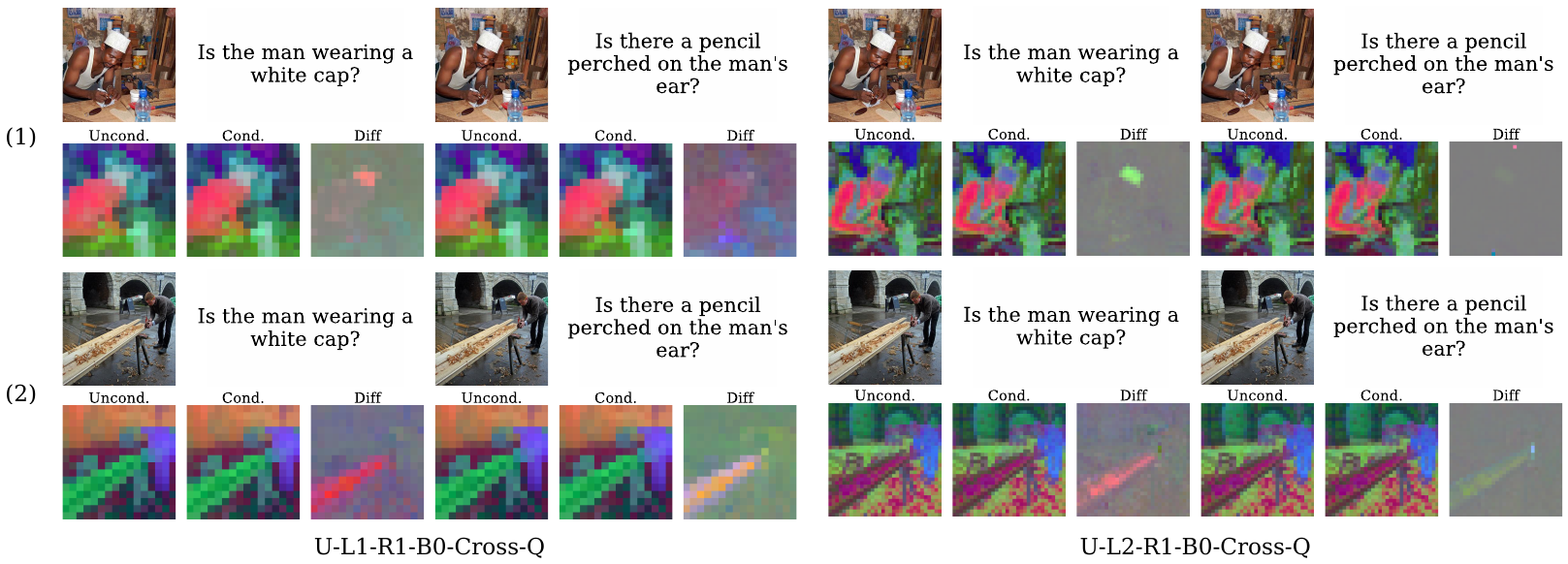}
    \includegraphics[width=1\linewidth]{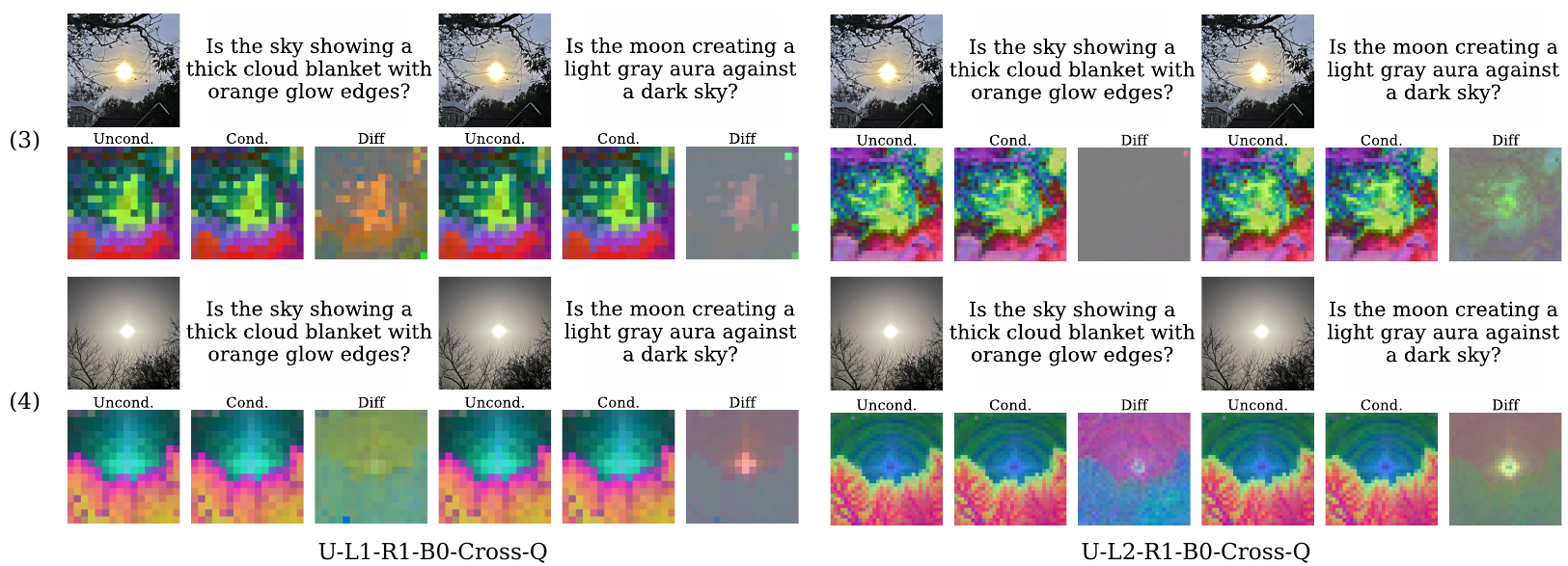}
    \caption{\textbf{More Visualizations of Question-Conditioned Features.} Please zoom in to see smaller regions of focus (e.g., pencil in second row, rightmost column of \texttt{U-L2-R1-B0-Cross-Q})}
    \label{fig:question_cond_pca_suppl}
\end{figure}

In this section, we provide two more instances of how providing questions as input prompts to the diffusion model enables focus on the relevant regions. We show these in Fig.~\ref{fig:question_cond_pca_suppl}. To understand how question conditioning may change across resolutions, we also provide comparisons for both \texttt{U-L1-R1-B0-Cross-Q} and \texttt{U-L2-R1-B0-Cross-Q} features. 

Notably, we observe that for some questions, high-resolution features can better focus on relevant regions. This is observed for the second image in the first pair of images. Specifically, for the question about the ``pencil on the man’s ear,'' we find that \texttt{U-L2-R1-B0-Cross-Q} features highlight the pencil, while \texttt{U-L1-R1-B0-Cross-Q} features instead focus on the wooden board. We also find that when the question is not aligned, the model can increase focus on the general foreground as seen in rows 3 and 4 for the \texttt{U-L1-R1-B0-Cross-Q} features. Alternatively, the model may also have a more diffuse focus or no change in focus as shown in rows 1 and 2 for the \texttt{U-L2-R1-B0-Cross-Q} features.

These examples further demonstrate the complementary information in features across different layers and resolutions. While our analysis focused on a single set of question-conditioned features, future work could explore more advanced ensembling techniques to improve model performance.

\section{Investigating SDXL Architecture}

\begin{table*}[h]
\centering
\caption{{Performance Comparison of Uncond. Diffusion Features at T=50.}}
\label{tab:sd_vs_sdxl}
\setlength{\tabcolsep}{5pt}
\renewcommand{\arraystretch}{1.2}
\begin{tabular}{@{}lcccc@{}}
\toprule
\textbf{Model} & \textbf{ROUGE-L} & \textbf{CIDEr} & \textbf{BLEU @ 4} & \textbf{SPICE} \\
\midrule
Stable-Diffusion-2.1-base (512)  & 36.25 & 26.20 & \textbf{11.54} & \textbf{16.21} \\
Stable-Diffusion-XL-base (512)   & \textbf{37.24} & \textbf{28.30} & 11.41 & 15.53 \\
\bottomrule
\end{tabular}
\end{table*}

In this section, we investigate whether some of our key trends extend to the SDXL architecture. Namely, we focus on how well SDXL features compare to SD2.1 features and use the same protocol as Sec~\ref{subsec:leakage}. We extract features from \texttt{up-level0-repeat0-vit-block7-out} from SDXL (based on analysis done in~\cite{meng2024diffusionmodelactivationsevaluated}). For this analysis, only the projection layer is trained, while both the vision encoder and LLM remain frozen. Our results are shown in Table~\ref{tab:sd_vs_sdxl}. We observe that there are slight improvements in ROUGE and CIDEr scores when using SDXL, but SD2.1 outperforms SDXL on BLEU and SPICE. 

\begin{table*}[h]
\centering
\caption{{Measuing Leakage Effects with SDXL Features via Mismatched Setting}}
\label{tab:sdxl_leakage}
\setlength{\tabcolsep}{5pt}
\renewcommand{\arraystretch}{1.2}
\begin{tabular}{@{}lcccc@{}}
\toprule
\textbf{Model} & \textbf{ROUGE-L} & \textbf{CIDEr} & \textbf{BLEU @ 4} & \textbf{SPICE} \\
\midrule

\midrule
SDXL-GT Captions ($s = 1$)  & 29.25 & 7.32 & 4.15 & 4.64
\\ % Row 15
SDXL-GT Captions ($s = 1.5$) & 36.03 & 21.39 & 8.99 & 9.78
\\ % Row 27
SDXL-GT Captions ($s = 1.5$) w/ 30\% caption dropout & 30.96 & 10.39 & 5.30 & 6.04 
\\ % Row 27
SDXL-GT Captions ($s = 4$) & 49.58 & 63.48 & 20.32 & 21.43
\\ % Row 27
\bottomrule
\end{tabular}
\end{table*}

We also examine whether SDXL features exhibit similar leakage effects as SD2.1 features and perform the same comparison as in Sec~\ref{subsec:leakage} but using SDXL features. Specifically, we use the ``Mismatched'' setting where we pass the ground-truth caption as input to SDXL along with a randomly chosen incorrect image. As shown in Table~\ref{tab:sdxl_leakage}, we observe that SDXL exhibits even stronger text-dependence. When trained with a guidance scale of 4, the model can more easily extract text-prompt features to reconstruct this caption even when the image is completely different. We find that adding dropout is still able to mitigate this effect.

\end{document}